\documentclass[final]{cvpr}
\usepackage{times}
\usepackage{epsfig}
\usepackage{amsmath}
\usepackage{amssymb}

\usepackage{booktabs}
\usepackage{tablefootnote}
\usepackage{multirow}
\usepackage{nicefrac}       % compact symbols for 1/2, etc.

\usepackage{cprotect}

\usepackage[pagebackref=true,breaklinks=true,colorlinks,bookmarks=false]{hyperref}

\def\cvprPaperID{244} % *** Enter the CVPR Paper ID here
\def\confYear{CVPR 2021}

\begin{document}

%%%%%%%%% TITLE
\title{Understanding Object Dynamics for Interactive Image-to-Video Synthesis}

\author{
Andreas Blattmann  \qquad Timo Milbich \qquad Michael Dorkenwald  \qquad Bj{\"o}rn Ommer\\
Interdisciplinary Center for Scientific Computing, HCI \\
Heidelberg University, Germany\\
}

\maketitle

%%%%%%%%% ABSTRACT
\begin{abstract}
        What would be the effect of locally poking a static scene? We present an approach that learns naturally-looking global articulations caused by a local manipulation at a pixel level. Training requires only videos of moving objects but no information of the underlying manipulation of the physical scene. Our generative model learns to infer natural object dynamics as a response to user interaction and learns about the interrelations between different object body regions. Given a static image of an object and a local poking of a pixel, the approach then predicts how the object would deform over time. In contrast to existing work on video prediction, we do not synthesize arbitrary realistic videos but enable local interactive control of the deformation. Our model is not restricted to particular object categories and can transfer dynamics onto novel unseen object instances. Extensive experiments on diverse objects demonstrate the effectiveness of our approach compared to common video prediction frameworks. Project page is available at \small{\url{https://bit.ly/3cxfA2L}}.
\end{abstract}

%%%%%%%%% BODY TEXT
\section{Introduction}
% Humans learn about the world by interaction with their immediate environment and subsequently observing the various responses to their actions.
From infancy on we learn about the world by manipulating our immediate environment and subsequently observing the resulting diverse reactions to our interaction. Particularly in early years, poking, pulling, and pushing the objects around us is our main source for learning about their integral parts, their interplay, articulation and dynamics. Consider children playing with a plant. They eventually comprehend how subtle touches only affect individual leaves, while increasingly forceful interactions may affect larger and larger constituents, thus finally learning about the entirety of the dynamics related to various kinds of interactions. Moreover, they learn to generalize these dynamics across similar objects, thus becoming able to predict the reaction of a novel object to their manipulation.

Training artificial visual systems to gain a similar understanding of the distinct characteristics of object articulation and its distinct dynamics is a major line of computer vision research. In the realm of still images the interplay between object shape and appearance has been extensively studied, even allowing for controlled, global~\cite{mathieu,ma2017disentangled,two_step,vunet,robust} and local~\cite{partbased,Zanfir_2018_CVPR} manipulation. Existing work on object dynamics, however, so far is addressed by either extrapolations of observed object motion \cite{2018epva,jin2020,svg,2018savp,2018epva,2019svrnn,vrnn-hier,villegas19} or only coarse control of predefined attributes such as explicit action labels \cite{Yang_2018_ECCV} and imitation of previously observed holistic motion \cite{aberman2020unpaired,hbugen,Blattmann_2021_CVPR}. Directly controlling and, even further, interacting with objects on a local level however, so far, is a novel enterprise.
\begin{figure}
\begin{center}
\includegraphics[width=.5\textwidth]{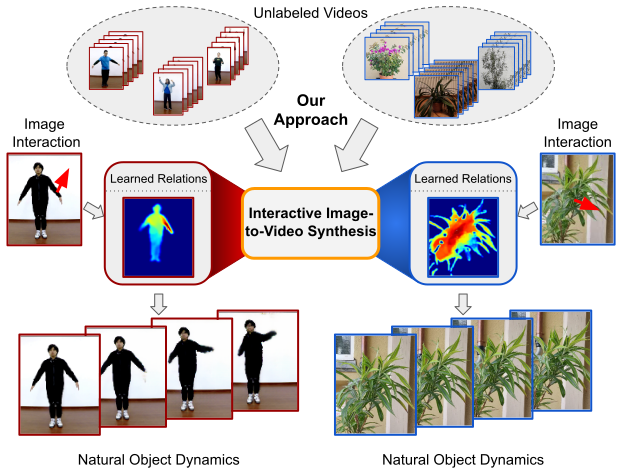}
\label{fig:fpf}
\caption{Our approach for interactive image-to-video synthesis learns to understand the relations between the distinct body parts of articulated objects from unlabeled video data, thus enabling synthesis of videos showing natural object dynamics as responses to local interactions.}
\end{center}
\vspace{-6mm}
\end{figure}
Teaching visual systems to understand the complex dynamics of objects both arising by explicit manipulations of individual parts and to predict and analyze the behavior~\cite{magnification_dorkenwald, uBAM_Biagio, LSTM_biagio} of the remainder of the object is an exceptionally challenging task. Similarly to a child in the example above, such systems need to know about the natural interrelations of different parts of an object~\cite{contour_shape_eccv12}. Moreover, they have to learn how these parts are related by their dynamics to plausibly synthesize temporal object articulation as a response to our interactions.

In this paper, we present a generative model for \textit{interactive image-to-video synthesis} which learns such a fine-grained understanding of object dynamics and, thus, is able to synthesize video sequences that exhibit natural responses to local user interactions with images on \textit{pixel-level}. Using intuitions from physics~\cite{strogatz}, we derive a hierarchical recurrent model dedicated to model complex, fine-grained object dynamics. Without making assumptions on objects we learn to interact with and no ground-truth interactions provided, we learn our model from video sequences only.

We evaluate our model on four video datasets comprising the highly-articulated object categories of humans and plants. Our experiments demonstrate the capabilities of our proposed approach to allow for fine-grained user interaction. Further, we prove the plausibility of our generated object dynamics by comparison to state-of-the-art video prediction methods in terms of visual and temporal quality. Figure~\ref{fig:fpf} provides an overview over the capabilities of our model.
%-------------------------------------------------------------------------
\section{Related Work}
\textbf{Video Synthesis.}
Video Synthesis involves a wide range of tasks including video-to-video translation~\cite{vid2vid}, image animation \cite{Wiles_2018_ECCV,Siarohin_2019_CVPR,first_order_2019}, frame interpolation \cite{Niklaus_2018_CVPR,xue2019video,liu2019cyclicgen,Bao_2019_CVPR,Niklaus_2020_CVPR}, unconditional video generation~\cite{torralba2016,mocogan,clark2020adversarial,iccv17_unsup_video} and video prediction.
Given a set of context frames, video prediction methods aim to predict realistic future frames either deterministically~\cite{2018epva,villegas17,van2018relational,Wu_2020_CVPR} or stochastically~\cite{cdna2016,svg,villegas17,2018savp,2018epva,sdcnet18,2019svrnn,vrnn-hier,villegas19,Kumar2020VideoFlow:}. A substantial number of methods ground on image warping techniques~\cite{vondrick5,voxel_flow,Gao_2019_ICCV}, leading to high-quality short term predictions, while struggling with longer sequence lengths.
% A substantial number of works grounds on image warping techniques~\cite{vondrick5,voxel_flow,Gao_2019_ICCV}, as these in general yield high-quality short term predictions. However, those methods do not model temporal relations between image frames, but warp a given input state instead, causing them to not generalize well to longer sequence lengths.
To avoid such issues, many works use autoregressive models. Due to consistently increasing compute capabilities, recent methods aim to achieve this task via directly maximizing likelihood in the pixel space, using large scale architectures such as normalizing flows~\cite{glow,Kumar2020VideoFlow:} or pixel-level transformers~\cite{NIPS2016_b1301141,Weissenborn2020Scaling}. As such methods introduce excessive computational costs and are slow during inference, most existing methods rely on RNN-based methods, acting autoregressively in the pixel-space~\cite{matthieu16,2018savp,LuHirSch17,svg,villegas17,vrnn-hier} or on intermediate representation such as optical flow~\cite{liang_2017,Li_2018_ECCV,Pan_2019_CVPR}. However, since they have no means for direct interactions with a depicted object, but instead rely on observing past frames, these methods model dynamics in a holistic manner. By modelling dynamics entirely in the latent space, more recent approaches take a step towards a deeper understanding of dynamics~\cite{2019svrnn,Franceschi2020,Dorkenwald_2021_CVPR} and can be used to factorize content from dynamics~\cite{Franceschi2020}, which are nonetheless modeled holistically. In contrast, our model has to infer plausible motion based on local interactions and, thus, understands dynamics in a more fine-grained way.\\
\noindent
\textbf{Controllable Synthesis of Object Dynamics.}
Since it requires to understand the interplay between their distinct parts, controlling the dynamics of articulated objects is a highly challenging task. Davis et al.~\cite{davis2015} resort to modeling rigid objects as spring-mass systems and animate still image frames by evaluating the resulting motion equations. However, due to these restricting assumptions, their method is only applicable for small deviations around a rest state, thus unable model complex dynamics.

To reduce complexity, existing learning based approaches often focus on modelling human dynamics using low-dimensional, parametric representations such as keypoints~\cite{aberman2020unpaired,Yang_2018_ECCV,Blattmann_2021_CVPR}, thus preventing universal applicability. Moreover, as these approaches are either based on explicit action labels or require motion sequences as input, they cannot be applied to controlling single body parts.  When intending to similarly obtain control over object dynamics in the pixel domain, previous methods use ground truth annotations such as holistic motion trajectories~\cite{cdna2016,2018savp,Dorkenwald_2021_CVPR} for simple object classes without articulation~\cite{bair}. Hao et al.~\cite{controllable_image} step towards locally controlling the video generation process by predicting a single next images based on a given input frame and sets of sparse flow vectors. Their proposed approach, however,
%as they solely infer a flow field from these inputs and warp the given image, they are restricted to holistic global object motions. Furthermore, the approach
requires multiple flow vectors for each individual frame of a sequence to be predicted, thus preventing localized, fine-grained control. Avoiding such flaws and indeed allowing for localized control, our approach introduces a latent dynamics model, which is able to model complex, articulated motion based on an interaction at a \textit{single} pixel.
\begin{figure*}[t]
    \centering
    \includegraphics[width=\textwidth]{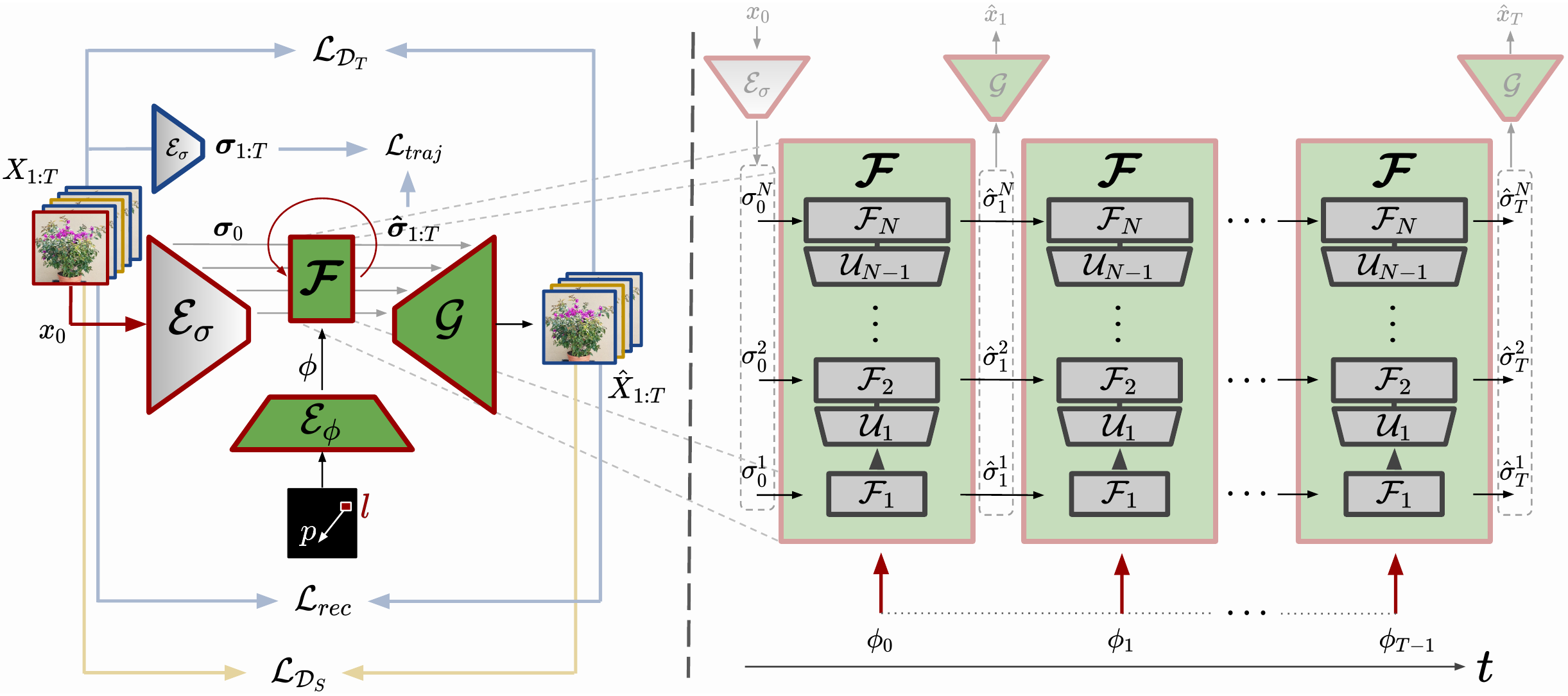}
    \caption{\textit{Left: }Our framework for interactive image-to-video synthesis during training. \textit{Right:} Our proposed hierarchical latent model $\boldsymbol{\mathcal{F}}$ for synthesizing dynamics, consisting of a hierarchy of individual RNNs $\mathcal{F}_n$, each of which operates on a different spatial feature level of the UNet defined by the pretrained encoder $\mathcal{E}_\sigma$ and the decoder $\mathcal{G}$. Given the initial object state $\boldsymbol{\sigma}_0 = [\mathcal{E}_\sigma(x_0)^1,...,\mathcal{E}_\sigma(x_0)^N]$, $\boldsymbol{\mathcal{F}}$ predicts the next state $\boldsymbol{\hat{\sigma}}_{i+1} = [\hat{\sigma}_{i+1}^1,...,\hat{\sigma}_{i+1}^N]$ based on its current state $\hat{\sigma}_i$ and the latent interaction $\phi_i = \mathcal{E}_\phi(p,l)$ at the corresponding time step. The decoder $\mathcal{G}$ finally visualizes each predicted object state $\hat{\sigma}_i$ in an image frame $\hat{x}_i$.}
    \label{fig:method1}
    % \vspace{-4mm}
\end{figure*}
\section{Interactive Image-to-Video Synthesis}
\label{sec:approach}
Given an image frame $x_0 \in \mathbb{R}^{H \times W \times 3}$, our goal is to interact with the depicted objects therein, i.e. we want to initiate a poke $p \in \mathbb{R}^2$ which represents a shift of a single location $l \in \mathbb{N}^2$ within $x_0$ to its new target location. Moreover, such pokes should also influence the remainder of the image in a natural, physically plausible way.
% dictated by the depicted object to correctly model.\\

Inferring the implications of local interactions upon the entire object requires a detailed understanding of its articulation, and, thus, of the interrelation between its various parts. Consequently, we require a structured and concise representation of the image $x_0$ and a given interaction. To this end, we introduce two encoding functions: an object encoder $\mathcal{E}_\sigma$ mapping images $x$ onto a latent object state $\sigma = \mathcal{E}_\sigma(x)$, e.g. describing current object pose and appearance, and the encoder $\mathcal{E}_\phi$ translating the target location defined by $p$ and $l$ to a latent interaction $\phi = \mathcal{E}_\phi(p,l)$ now affecting the initially observed object state $\sigma_0 = \mathcal{E}_\sigma(x_0)$.

Eventually, we want to synthesize a video sequence depicting the response arising from our interaction with the image $x_0$, represented by means of $\sigma_0$ and $\phi$. Commonly, such conditional video generation tasks are formulated by means of learning a video generator $\mathcal{G} : (\sigma_0,\phi) \rightarrow X_{1:T} = \{x_1, ..., x_T\}$~\cite{2018savp,vrnn-hier}. Thus, $\mathcal{G}$ would both model object dynamics and infer their visualization in the RGB space. However, every object class has its distinct, potentially very complex dynamics which - affecting the \textit{entire} object - must be inferred from a localized poke shifting a \textit{single pixel}. Consequently, a model for interactive image-to-video synthesis has to understand the complex implications of the poke for the remaining object parts and, thus, requires to model these dynamics in sufficiently fine-grained and flexible way. Therefore, we introduce a dedicated object dynamics model inferring a trajectory of object states $[\sigma_0, \sigma_1,\dots,\sigma_T]$ representing an object's response to $\phi$ within an object state space $\Omega$. As a result, $\mathcal{G}$ only needs to generate the individual images $x_i = \mathcal{G}(\sigma_i)$, thus decomposing the overall image-to-video synthesis problem.
\begin{figure*}[t]
\begin{center}
\includegraphics[width=\textwidth]{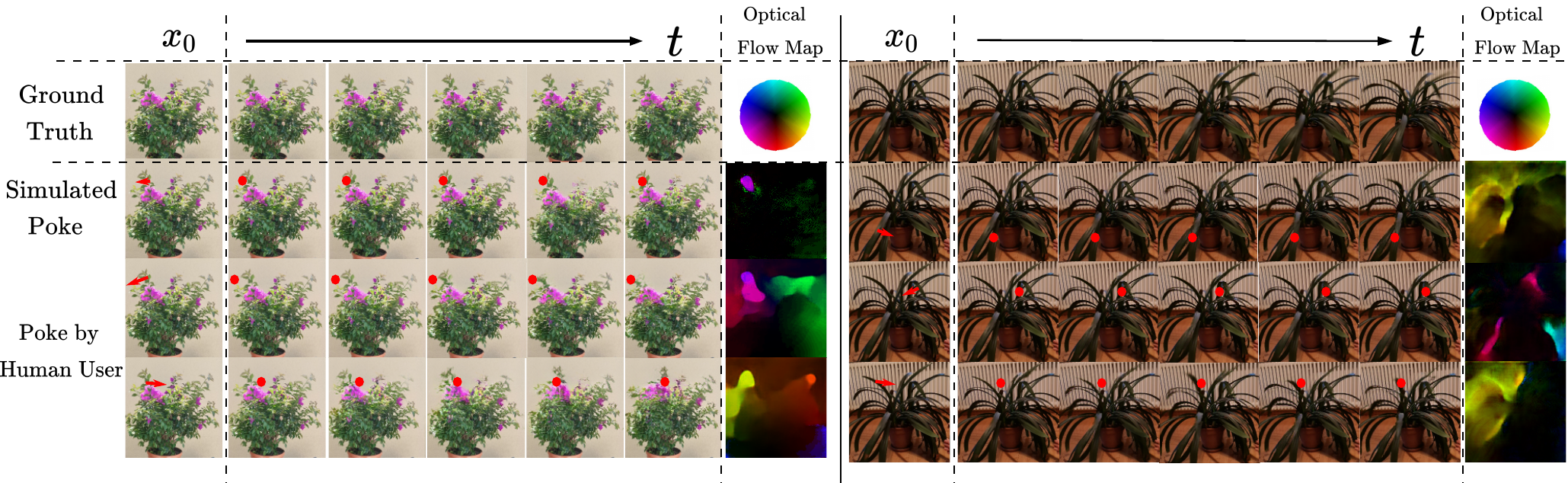}
\caption{Visualization of the videos generated by our model for two distinct plants within our self-recorded PokingPlants-Dataset: The first row depicts a ground truth sequence from the test set. The second row contains a simulated poke (red arrow) based on this ground truth sequence, using the procedure described in Section \ref{sec:training}. The last two rows show results of the model to pokes from human users. In the first column, the red arrow indicates the interaction and, thus, also the resulting target location, which is indicated as a red dot in the remaining columns. As the motions within the Poking-Plants dataset are sometimes subtle and not straightforward to detect, we also visualize the visual flow field, which was estimated based on the synthesized videos. We encourage the reader to also view the corresponding video results in the supplemental and on our project page \url{https://bit.ly/3cxfA2L}.}
\label{fig:vis_plants}
\end{center}
\vspace{-5mm}
\end{figure*}
\subsection{A Hierarchical Model for Object Dynamics}
In physics, one would typically model the trajectory of object states $\sigma(t)$ as a dynamical system and, thus, describe it as an ordinary differential equation (ODE)~\cite{strogatz, chang2018}

\begin{equation}
    \Dot{\sigma}(t)= f(\sigma(t), \phi(t)) \, , \, \sigma(0) = \sigma_0 \, ,
    \label{eq:dyn_sys}
\end{equation}
with $f$ the - in our case unknown - evolution function, $\Dot{\sigma}$ its first time derivative, and $\phi(t) = \phi~,~\forall t \in [0,T]$ the latent external interaction obtained from the poke. Recent work proposes to describe $f$ with fixed model assumptions, such as as an oscillatory system~\cite{davis2015}. While this may hold in some cases, it greatly restricts applications to arbitrary, highly-articulated object categories. Avoiding such strong assumptions, the only viable solution is to learn a flexible prediction function $\mathcal{F}$ representing the dynamics in Eq.~\eqref{eq:dyn_sys}. Consequently, we base $\mathcal{F}$ on recurrent neural network models~\cite{lstm, gru} which can be interpreted as a discrete, first-order approximation\footnote{Order here means order of time derivative. For more information regarding this correspondence between ODEs and RNNs, see~\cite{Haber_2017,chang2018,changhaber_2018,niu2019recurrent}.} to Eq.~\eqref{eq:dyn_sys} at time steps $i \in [0,T-1]$
\begin{equation}
    \mathcal{F}(\sigma_i, \phi_i) = \sigma_{i+1} = \sigma_i + h \cdot f_a(\sigma_i, \phi_i) \, ,
    \label{eq:euler_fw}
\end{equation}
with $h$ being the step size between two consecutive predicted states $\sigma_i$ and $\sigma_{i+1}$ and $f_a$ an approximation to the derivative at $\sigma_i$~\cite{chang2018, niu2019recurrent}. However, dynamics of objects can be arbitrarily complex and subtle such as leaves of a tree or plant fluttering in the wind. In such cases, the underlying evolution function $f$ is expected to be similarly complex and, thus, involving only first-order derivatives when modelling Eq.~\eqref{eq:dyn_sys} may not be sufficient. Instead, capturing also such high-frequency details actually calls for higher-order terms.

In fact, one can model an $N$-th order evolution function in terms of $N$ first order ODEs, by introducing a hierarchy $\boldsymbol{\sigma} = [\sigma^1,...,\sigma^N]\, , \, \sigma^1 = \sigma$ of state variables, the $n$-th element of which is proportional to the $(n-1)$-th order of discrete time derivative of the original variable $\sigma$~\cite{hale2009ordinary,niu2019recurrent}.
Consequently, as first order ODEs can be well approximated with Eq.~\eqref{eq:euler_fw}, we can extend $\mathcal{F}$ to a hierarchy of predictors $\boldsymbol{\sigma} = \boldsymbol{\mathcal{F}} = [\mathcal{F}_1, ..., \mathcal{F}_N]$ by using a sequence of $N$~RNNs,
\begin{equation}
    \sigma^n_{i+1} = \mathcal{F}_n(\sigma_{i}^n,\sigma_{i+1}^{n-1}) \, , \sigma^n_0 = \sigma_0
    \label{eq:hier_n_pre}
\end{equation}
each operating on the input of its predecessor, except for the lowest level $\mathcal{F}_1$, which predicts the coarsest approximation of the object states based on $\phi_i$ as
\begin{equation}
{\sigma}^1_{i+1} = \mathcal{F}_1(\sigma^1_{i},\phi_i) \, , \, \sigma^1_0 = \sigma_0 \, .
\label{eq:hier_1}
\end{equation}
A derivation of our this hierarchy is given in the Appendix~\ref{sec:deriv}.
However, while $\boldsymbol{\mathcal{F}}$ is able to approximate higher-order derivatives up to order $N$, thus being able to model fine-grained dynamics, we need to make sure that our decoder $\mathcal{G}$ actually captures these details when generating the individual image frames $x_i$.

Recent work on image synthesis indicates that standard decoder architectures fed with latent encodings only on the bottleneck-level, are prone to missing out on subtle image details~\cite{stylegan,spade}, such as those arising from high motion frequencies. Instead, providing a decoder with latent information at each spatial scale has proven to be more powerful~\cite{unet}. Hence, we model $\mathcal{G}$ to be the decoder of a hierarchical image-to-sequence UNet with the individual predictors $\mathcal{F}_n$ operating on the different spatial feature levels of $\mathcal{G}$. To maintain the hierarchical structure of $\boldsymbol{\mathcal{F}}$, we compensate for the resulting mismatch between the dimensionality of $\sigma_i^{n-1}$ and $\sigma_i^{n}$ by means of upsampling functions $\mathcal{U}_n$. Finally, to fully exploit the power of a UNet structure, we similarly model the encoder $\mathcal{E}_\sigma$ to yield a hierarchical object state $\boldsymbol{\sigma}_0 = [\mathcal{E}_\sigma(x_0)^1,...,\mathcal{E}_\sigma(x_0)^N]$, which is the initial state to $\boldsymbol{\mathcal{F}}$. Thus, Eq.~\eqref{eq:hier_n_pre} becomes
\begin{equation}
   \sigma^n_{i+1} = \mathcal{F}_n(\sigma_{i}^n,\mathcal{U}_{n-1}(\sigma_{i+1}^{n-1})) \, , \sigma^n_0 = \mathcal{E}_\sigma(x_0)^n
    \label{hier_n}
\end{equation}
with $\sigma^n_i$ being the predicted object state at feature level $n \in [2,N]$ and time step $i$. Hence, at each time step we obtain a hierarchy of $N$ latent object states $\boldsymbol{\sigma}_i = [\sigma^1_i,...,\sigma^N_i]$ on different spatial scales, which are the basis for synthesizing image frames $x_i$ by $\mathcal{G}$. Our full hierarchical predictor $\boldsymbol{\mathcal{F}}$ and its linkage to the overall architecture are shown in the right and left parts of Figure \ref{fig:method1}.

As our proposed model jointly adds more subtle details in the temporal \textit{and} spatial domain, it accurately captures complex dynamics arising from interactions and simultaneously  displays them in the image space.
\begin{figure*}[t]
\begin{center}
\includegraphics[width=\textwidth]{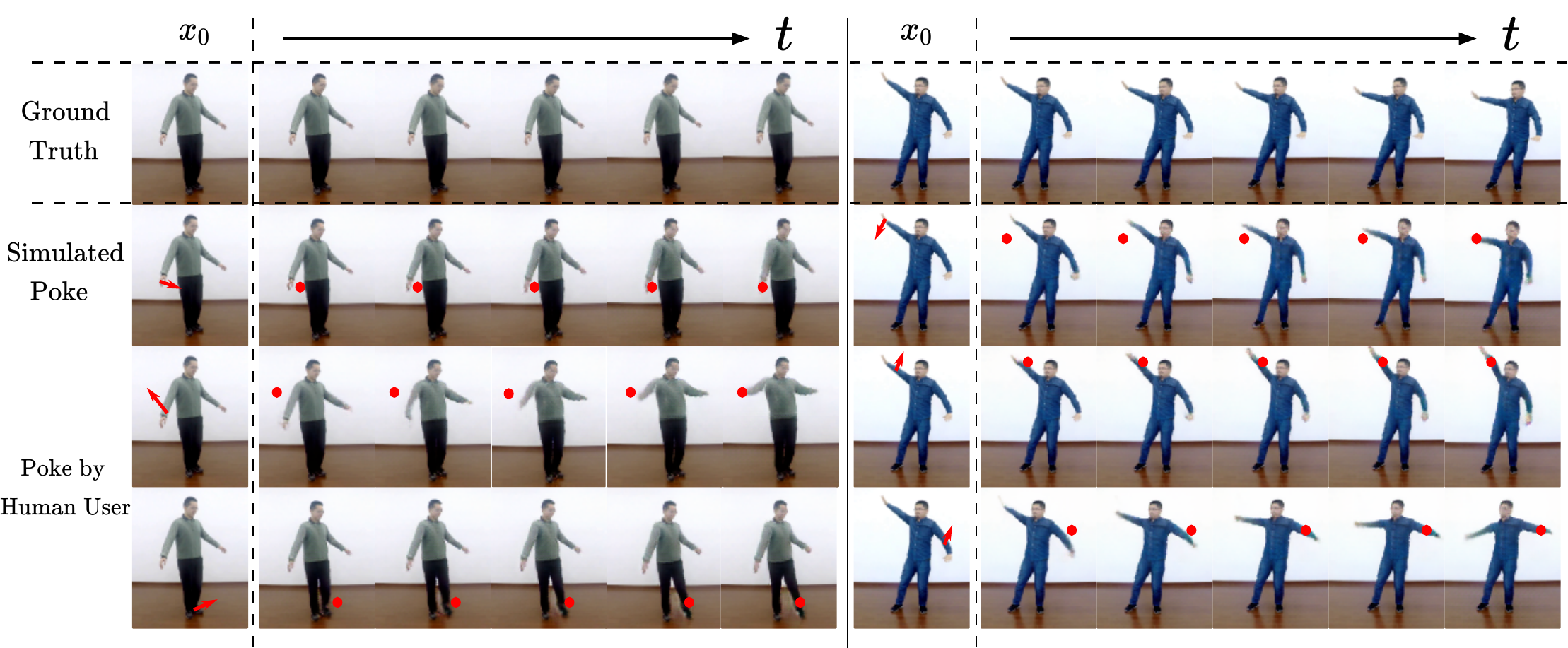}
\caption{Visualization of the videos generated by our model for two actors within the test set of the iPER~\cite{iper} dataset: The first row depicts the ground truth sequence. The second row contains a simulated poke (red arrow) based on this ground truth sequence, using the procedure described in Section \ref{sec:training}. The last two rows show results of the model to pokes from human users. In the first column, the red arrow indicates the interaction and, thus, also the resulting target location, which is indicated as a red dot in the remaining columns. We encourage the reader to also view the accompanying video results in the supplemental and on our project page \url{https://bit.ly/3cxfA2L}.
% \textsuperscript{\ref{project_page}}.
}
\label{fig:vis_iper}
\end{center}
\vspace{-6mm}
\end{figure*}
\subsection{Learning Dynamics from Poking}
\label{sec:training}
To learn our proposed model for interactive image-to-video synthesis, we would ideally have ground-truth interactions provided, i.e. actual pokes and videos of their immediate impact on the depicted object in $x_0$. However, gathering such training interactions is tedious and in some cases not possible at all, particularly hindering universal applicability. Consequently, we are limited to cheaply available video sequences and need to infer the supervision for interactions automatically.

While at inference a poke $p$ represents an \textit{intentional} shift of an image pixel in $x_0$ at location $l$, at training we only require to observe responses of an object to \textit{some} shifts, as long as the inference task is well and naturally approximated. To this end, we make use of dense optical flow displacement maps~\cite{flownet2} $D \in \mathbb{R}^{H \times W \times 2}$ between images $x_0$ and $x_T$ of our training videos, i.e. their initial and last frames. Simulating training pokes in $x_0$ then corresponds to sampling pixel displacements $p = (D_{l_1,l_2,1},D_{l_1,l_2,2})$ from $D$, with the video sequence $X_{1:T} = \{x_i\}_{i=1}^T$ being a natural response to $p$. Using this, we train our interaction conditioned generative model to minimize the mismatch between the individual predicted image frames $\hat{x}_i = \mathcal{G}(\boldsymbol{\mathcal{F}}(\boldsymbol{\sigma}_{i-1},\phi_{i-1}))$ and those of the video sequence as measured by the perceptual distance~\cite{Johnson2016Perceptual}
% for reconstructing $X_{1:T}$ applying the widely employed perceptual loss~\cite{}\red{(cite!)}
%
\begin{equation}
        \mathcal{L}_{rec} = \sum_{i=1}^{T} \sum_{k=1}^{K} \Vert \Phi_k(x_{i}) - \Phi_k(\hat{x}_{i}) \Vert_1 \, ,
        \label{eq:vid_rec}
\end{equation}
%
% where $\hat{x}_i$ denotes an individual reconstructed image frame and $\Phi_k$ denotes the $k$-th layer of a pre-trained VGG \cite{vgg} feature extractor.
where $\Phi_k$ denotes the $k$-th layer of a pre-trained VGG \cite{vgg} feature extractor.

However, due to the direct dependence of $\mathcal{G}$ on $\boldsymbol{\sigma}_i$, during end-to-end training the state space $\Omega$ is continuously changing. Thus, learning object dynamics by means of our hierarchical predictor $\boldsymbol{\mathcal{F}}$ is aggravated. To alleviate this issue, we propose to first learn a fixed object space state $\Omega$ by pretraining $\mathcal{E}_\sigma$ and $\mathcal{G}$ to reconstruct individual image frames. Training $\boldsymbol{\mathcal{F}}$ on $\Omega$ to capture dynamics depicted in $\{x_i\}_{i=0}^T$ is then performed by predicting states $\boldsymbol{\hat{\sigma}}_i = \boldsymbol{\mathcal{F}}(\boldsymbol{\hat{\sigma}}_{i-1}, \phi_i)$ approaching the individual states $\boldsymbol{\sigma}_i = [\mathcal{E}_\sigma(x_i)^1,...,\mathcal{E}_\sigma(x_i)^N]$ of the target trajectory, while simultaneously fine-tuning $\mathcal{G}$ to compensate for prediction inaccuracy, using
\begin{equation}
    \mathcal{L}_{traj} = \sum_{i=1}^{T} \sum_{n=1}^{N} \Vert \mathcal{E}_\sigma(x_i)^n - \hat{\sigma}^n_i\Vert_2 \, .
    \label{eq:latent_loss}
\end{equation}
Finally, to improve the synthesis quality we follow previous work \cite{clark2020adversarial,vid2vid} by training discriminators $\mathcal{D}_S$ on frame level and $\mathcal{D}_T$ on the temporal level, resulting in loss functions $\mathcal{L}_{\mathcal{D}_S}$ and $\mathcal{L}_{\mathcal{D}_T}$. Our overall optimization objective then reads
\begin{equation}
    \mathcal{L} = \mathcal{L}_{rec} + \lambda_{traj} \cdot \mathcal{L}_{traj} + \lambda_{S} \cdot \mathcal{L}_{\mathcal{D}_S} \,
    + \lambda_{\mathcal{D}_T} \cdot \mathcal{L}_{D_T}
\end{equation}
with hyperparameters $\lambda_{traj}$, $\lambda_{S}$ and $\lambda_{T}$. The overall procedure for learning our network is summarized in Fig. \ref{fig:method1}.

\begin{figure*}[t]
    \centering
    \includegraphics[width=\textwidth]{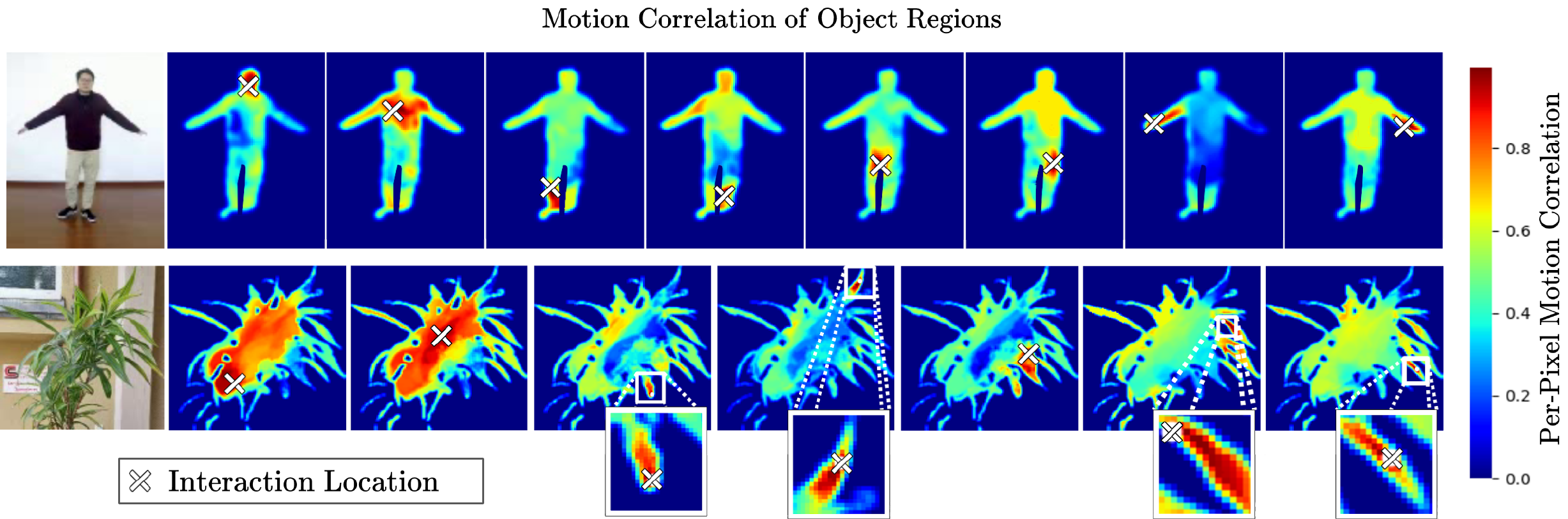}
    \caption{Understanding object structure: By performing 100 random interactions at the same location $l$ within a given image frame $x_0$ we obtain varying video sequences, from which we compute motion correlations for $l$ with all remaining pixels. By mapping these correlations to the pixel space, we visualize distinct object parts. For a detailed discussion, refer to Section~\ref{seq:eval_qual}.}
    \label{fig:part_vis}
    \vspace{-2mm}
\end{figure*}
\section{Experiments}
In this section we both qualitatively and quantitatively analyze our model for the task of interactive image-to-video synthesis. After providing implementation details, we illustrate qualitative results and analyze the learned object interrelations. Finally, we conduct a quantitative evaluation of both visual quality of our generated videos and the plausibility of the motion dynamics depicted within.
\subsection{Implementation Details and Datasets}
\label{ss:implementation}
\noindent
\textbf{Foreground-Background-Separation.}
While our model should learn to capture dynamics associated with objects initiated by interaction, interaction with areas corresponding to the background should be ignored. However, optical flow maps often exhibit spurious motion estimates in background areas, which may distort our model when considered as simulated object pokes for training. To suppress these cases we only consider flow vector exhibiting sufficient motion magnitude as valid pokes, while additionally training our model to ignore background pokes. More details are provided in Appendix~\ref{sec:model_train}. Examples for responses to background interactions can be found in the videos on our project page \url{https://bit.ly/3cxfA2L}.
\\
\textbf{Model Architecture and Training. } The individual units $\mathcal{F}_n$ of our hierarchical dynamics model are implemented as Gated Recurrent Units (GRU)~\cite{gru}. The depth $N$ of our hierarchy of predictors varies among datasets and videos resolutions. For the experiments shown in the main paper, we train our model to generate sequences of 10 frames and spatial size $128 \times 128$, if not specified otherwise. For training, we use ADAM \cite{adam} optimizer with parameters $\beta = (0.9, 0.99)$ and a learning rate of $10^{-4}$. The batch size during training is 10. The weighting factors for our final loss function are chosen as $\lambda_{traj}=0.1$, $\lambda_{\mathcal{D}_S} = 0.2$ and $\lambda_{\mathcal{D}_T} = 1$. More details regarding architecture, hyperparameters and training can be found in the Appendix~\ref{sec:implementation}. Additionally, alternate training procedures for different parameterizations of the poke and the latent interaction $\phi(t)$ are presented in Appendix~\ref{sec:exp_add}.
\\
\textbf{Datasets}
We evaluate the capabilities of our model to understand object dynamics on four datasets, comprising the highly-articulated object categories of humans and plants.

\textit{Poking-Plants (PP)} is a self-recorded dataset showing 27 video sequences of 13 different types of pot plants in motion. As the plants within the dataset have large variances in shape and texture, it is particularly challenging to learn a single dynamics model from those data. The dataset consists of overall 43k image frames, of which we use a fifth as test and the remainder as train data.

\textit{iPER} \cite{iper} is a human motion dataset containing 206 video sequences of 30 human actors of varying shape, height gender and clothing, each of which is recorded in 2 videos showing simple and complex movements. We follow the predefined train test split with a train set containing 180k frames and a test set consisting of 49k frames.

\textit{Tai-Chi-HD} \cite{first_order_2019} is a video collection of 280 in-the-wild Tai-Chi videos of spatial size $256 \times 256$ from youtube  which are subdivided into 252 train and 28 test videos, consisting of 211k and 54k image frames. It contains large variances in background and also camera movements, thus serving as an indicator for the real-world applicability of our model. As the motion between subsequent frame is often small, we temporally downsample the dataset by a factor of two.

\textit{Human3.6m \cite{h36m}} is large scale human motion dataset, containing video sequences of 7 human actors performing 17 distinct actions. Following previous work on video prediction \cite{2018epva,2019svrnn,Franceschi2020}, we centercrop and downsample the videos to 6.25 Hz and use actors S1,S5,S6,S7 and S8 for training  and actors S9 and S11 for testing.
\subsection{Qualitative Analysis}
\label{seq:eval_qual}
\begin{table}
    \centering
    \resizebox{.9\linewidth}{!}{
    \begin{tabular}{l||c|c|c|c}
    \toprule
    Dataset & PP & iPER  & Tai-Chi  & H3.6m  \\
    \hline
    SAVP \cite{2018savp} &  136.81 & 150.39  & 309.62 &  160.32  \\
    IVRNN \cite{vrnn-hier}  & 184.54 &  181.50 &  202.33 & 465.55  \\
    SRVP \cite{Franceschi2020}  & 283.90  & 245.13 & 238.71 & 174.65
    % \tablefootnote{Official pretrained model from https://github.com/edouardelasalles/srvp, conditioned on 8 frames}
    \\
    \hline
    Ours  & \textbf{89.67}  & \textbf{144.92}   & \textbf{171.74} &  \textbf{107.73} \\
    \hline
    \end{tabular}}
    \caption{Comparison with recent work from video prediction. Across datasets, our approach obtains considerably lower FVD-scores.}
    \label{tab:comp_vp}
    \vspace{-3mm}
\end{table}
\begin{table*}[t]
    \centering
    \resizebox{.76\textwidth}{!}{
    \begin{tabular}{l || c | c || c | c || c | c || c | c  }
    \toprule
    Dataset & \multicolumn{2}{c ||}{PP} & \multicolumn{2}{c|}{iPER \cite{iper}} & \multicolumn{2}{c|}{TaiChi \cite{first_order_2019}} & \multicolumn{2}{c}{Human3.6m \cite{h36m}}\\
    \hline
    Method & Ours & Hao et al. \cite{controllable_image}  &   Ours &  Hao et al. \cite{controllable_image} & Ours & Hao et al.  \cite{controllable_image} & Ours &  Hao et al. \cite{controllable_image} \\
    \hline
    \hline
     FVD $\downarrow$& \textbf{137.78} & 361.51 & \textbf{93.28} & 235.08    & \textbf{152.76} & 341.79  & \textbf{100.36} & 259.92 \\
    LPIPS $\downarrow$ & \textbf{0.10} & 0.16 & \textbf{0.06} & 0.11 & \textbf{0.12} & \textbf{0.12} & \textbf{0.08} & 0.10 \\
    PSNR $\uparrow$& 20.68 & \textbf{21.28} & \textbf{23.99} & 21.09 & \textbf{20.58} & 20.41  & \textbf{23.71} & 22.81 \\
    SSIM $\uparrow$& \textbf{0.78} & 0.72 & \textbf{0.91} & 0.88 & 0.75 & \textbf{0.78}  & 0.92 & \textbf{0.93} \\
    \hline
    \end{tabular}}
    \caption{Comparison on controlled video synthesis. We compare with Hao et al. \cite{controllable_image} which we consider the closest previous work to ours.}
    \label{tab:compare_ci}
    \vspace{-4mm}
\end{table*}
\noindent
\textbf{Interactive Image-to-Video-Synthesis.}~We now demonstrate the capabilities of our model for the task of interactive image-to-video synthesis. By poking a pixel within a source image $x_0$, i.e. defining the shift $p$ between the source location $l$ and the desired target location, we require our model to synthesize a video sequence showing the object part around $l$ approaching the defined target location. Moreover, the model should also infer plausible dynamics for the remaining object parts. Given $x_0$, we visualize the video sequences generated by our model for $i)$ simulated pokes obtained from optical flow estimates (see Sec.~\ref{sec:training}) from test videos and $ii)$ pokes initiated by human users. We compare them to the corresponding ground truth videos starting with $x_0$. For the object category of plants, we also show optical flow maps between the source frame and the last frame for each generated video, to visualize the overall motion therein.

Figure~\ref{fig:vis_plants} shows examples for two distinct plants from the PokingPlants dataset of very different appearance and shape. In the left example we interact with the same pixel location varying both poke directions and magnitudes. Our model correctly infers corresponding object dynamics. Note that the interaction location corresponds to a very small part of the plant, initiating both subtle and complex motions. This demonstrates the benefit of our hierarchical dynamics model. As shown by the visualized flow, the model also infers plausible movements for object parts not directly related to the interaction, thus indicating our model to also understand long term relations between distinct object regions. In the right example we observe similar results when interacting with different locations.

As our model does not make any assumptions on the depicted object to manipulate, we are not restricted to particular object classes. To this end we also consider the category of humans which is widely considered for various video synthesis tasks. Figure \ref{fig:vis_iper} shows results for two unseen persons from the iPER test set. Again our model infers natural object dynamics for both the interaction target location and the entire object. This highlights that our model is also able to understand highly structured dynamics and natural interrelations of object parts independent of object appearance. Analogous results for the Tai-Chi-HD and Human3.6m datasets are contained in the Appendix~\ref{sec:add_vis}.
\\
\textbf{Understanding Object Structure. }
We now analyze the interrelation of object parts learned by our model, i.e. how pixels depicting such parts correlate when interacting with a specific, fixed interaction location.

To this end, we perform $100$ random interactions for a given, fixed source frame $x_0$ at the fixed location $l = (l_1, l_2)$ of different magnitudes and directions. This results in varying videos $\{\hat{X}_k\}_{k=1}^{100}$, exhibiting distinct dynamics of the depicted object. To measure the correlation in motion of all pixels with respect to the interaction location, we first obtain their individual motion for each generated video using optical flow maps between $x_0$ and the last video frames. Next, for each pixel we compute the L2-distance in each video to the respective interaction poke based on a $[\text{magnitude},\text{angle}]$ representation, thus obtaining a $100$-dimensional difference vector. To measure the correlation for each pixel with $l$, we now compute the variance over each difference vector to obtain correlation maps. High correlation then corresponds to low variance in motion.

Figure~\ref{fig:part_vis} visualizes such correlations by using heatmaps for different interaction locations $l$ in the same source frame $x_0$ for both plants and humans. For humans, we obtain high correlations for the body parts around $l$, indicating our model to actually understand the human body structure. Considering the plant, we see that poking locations on its trunk (first two columns) intuitively results in highly similar movements of all those pixels close to the trunk. The individual leaves farther away are not that closely correlated, as they might perform oscillations in higher frequencies superposing the generally low-frequency movements of the trunk. When poking these leaves, however, mostly directly neighbouring pixels exhibit high correlation.
\subsection{Quantitative Evaluation}
To quantitatively demonstrate the ability of our approach to synthesize plausible object motion, we compare with the current state-of-the-art of RNN-based video prediction~\cite{2018savp,vrnn-hier,Franceschi2020}. For all competing methods we used the provided pretrained models, where available, or trained the models using official code. Moreover, to also evaluate the controllable aspect of our approach, we compare against Hao et al.~\cite{controllable_image}, an approach for controlled video synthesis by predicting independent single frames from sparse sets of flow vectors, which we consider the closest previous work to interactive image-to-video synthesis. We trained their method on our considered datasets using their provided code.
\\
\noindent
\textbf{Evaluation Metrics. }
For evaluation we utilize the following metrics. Additional details on all metrics are listed in Appendix~\ref{sec:eval}.

\textit{Motion Consistency. } The Fr\'echet-Video-Distance (FVD, lower-is-better) is the standard metric to evaluate video predictions tasks, as it is sensitive to the visual quality, as well as plausibility and consistency of motion dynamics of a synthesized video. Moreover it has been shown to correlate well with human judgement~\cite{fvd}. All FVD-scores reported hereafter are obtained from video sequences of length 10.

\textit{Prediction accuracy. } Our model is trained to understand the global impact of localized pokes on the overall object dynamics and to infer the resulting temporal object articulation.
To evaluate this ability, we report distance scores against the ground-truth using three commonly used frame-wise metrics, averaged over time: SSIM (higher-is-better) and PSNR (higher-is-better) directly compare the predicted and ground truth frames. Since they are based on L2-distances directly in pixel space, they are known disregard high-frequency image details, thus preferring blurry predictions. To compensate, we also report the LPIPS-score~\cite{lpips} (lower-is-better), comparing images based on distance between activations of pretrained deep neural network and, thus exhibit better correlation with human judgment.
\\
\textbf{Comparison to other methods. } We evaluate our approach against SAVP~\cite{2018savp}, IVRNN~\cite{vrnn-hier} and SRVP~\cite{Franceschi2020} which are RNN-based  state-of-the-art methods in video prediction based on FVD scores. We train these models to predict video sequences of length 10 given 2 context frames corresponding to the evaluation setting. Note, that our method generates sequences consisting of 10 frames based on a \textit{single} frame and a poke. Following common evaluation practice in video prediction~\cite{2018savp,vrnn-hier,Franceschi2020}, we predict sequences are $64 \times 64$ for all models, including ours. The results are shown in Tab.~\ref{tab:comp_vp}. Across datasets, our approach achieves significantly lower FVD-scores demonstrating the effectiveness of our approach to both infer and visualize plausible object dynamics. Particularly on the plants dataset our model performs significantly better than the competitors due to our proposed hierarchic model for capturing fine-grained dynamics.
%as uotherwise, the global motion would not be recovered for theentire object body, but only for the body part affected by the poke.

To quantitatively evaluate our ability for controlled video synthesis, we now compare against Hao et al.~\cite{controllable_image}. For them to predict videos $[x_0,\dots,x_T]$ of length $T$, we first sample $k$ flow vectors between $x_0$ and $x_1$. We then extend these initial vectors to discrete trajectories of length $T-1$ in the image space by tracking the initial points using consecutive flow vectors between $x_{i}$ and $x_{i+1}$. Video prediction is then performed by individually warping $x_0$ given the shift of its pixels to the intermediate locations at step $i$ in these trajectories. Following their protocol we set $k=5$. Note, that we only require \textit{a single interactive poke} of arbitrary length for synthesize videos. Tab.~\ref{tab:compare_ci} compares both approaches. We see that our model performs significantly better by a large margin, especially in FVD measuring the video quality and temporal consistency. This is explained by temporal object dynamics not even being considered in \cite{controllable_image}, in contrast to our dedicated hierarchical dynamics model. Further, their image warping-based approach typically results in blurry image predictions. Across datasets we outperform their method also in LPIPS scores indicating our proposed method to produce more visually compelling results.
\subsection{Ablation Studies}
Finally, we conduct ablation studies to analyze the effects of the individual components of our proposed method. To limit computational cost, we conduct all ablation experiments using videos of spatial size $64 \times 64$. \textit{Ours RNN} indicates our model trained with a common GRU consisting of three individual cells at the latent bottleneck of the UNet instead of our proposed hierarchy of predictors $\boldsymbol{\mathcal{F}}$. For fair evaluation with the baselines we also choose $N=3$ as depth of the hierarchy. Further, we also evaluate baselines without considering the loss term $\mathcal{L}_{traj}$ which compensates for the prediction inaccuracy of $\boldsymbol{\mathcal{F}}$ (\textit{Ours w/o $\mathcal{L}_{traj}$}) and without pretraining the object encoder $\mathcal{E}_\sigma$ (\textit{Ours single-stage}). Tab.~\ref{tab:Ablations} summarizes the ablation results, indicating the the benefits of each individual component across reported metrics and used datasets. We observe the largest impact when using our hierarchical predictor $\boldsymbol{\mathcal{F}}$, thus modelling higher-order terms in our dynamics model across all spatial feature scales. Looking at the remaining two ablations, we also see improvements in FVD-scores. This concludes that operating on a pretrained, consistent object state space $\Omega$ and subsequently accounting for the prediction inaccuracies, leads to significantly more stable learning. An additional ablation study on the effects of varying the highest order $N$ of modeled derivative by our method is provided in Appendix~\ref{sec:ode_deriv}.
\subsection{Additional Experiments}
We also conduct various alternate experiments, including the generalization of our model to unseen types of plants and different interpretations of the poke $p$, resulting in different capabilities of our model. These experiments are contained in the Appendix~\ref{sec:exp_add}, which also contains many videos further visualizing the corresponding results.
%------------------------------------------------------------------------
\section{Conclusion}
\vspace{-.5em}
In this work, we propose a generative model for \textit{interactive image-to-video synthesis} which learns to understand the dynamics of articulated objects by capturing the interplay between the distinct object parts, thus allowing to synthesize videos showing natural responses to localized user interactions. The model can be flexibly learned from unlabeled video data without limiting assumptions on object shape. Experiments on a range of datasets prove the plausibility of generated dynamics and indicate the model to produce visually compelling video sequences.
\begin{table}[t]
    \centering
    \resizebox{.45\textwidth}{!}{\begin{tabular}{l|c|c|c|c|c|c}
    \toprule
    Dataset & \multicolumn{3}{c|}{PP} & \multicolumn{3}{c}{iPER \cite{iper}} \\
    \hline
    Method & LPIPS $\downarrow$ & PSNR $\uparrow$ & FVD $\downarrow$ & LPIPS $\downarrow$ & PSNR $\uparrow$ & FVD $\downarrow$ \\
    \hline
    Ours RNN & 0.08 & 20.92  & 175.42 & 0.07 & 23.02  & 250.64 \\
    Ours w/o $\mathcal{L}_{traj}$ &  0.07 & 21.53  & 110.00  & 0.05 & 22.82 & 203.26 \\
    Ours (single) &  0.07 & 21.41  & 115.65  & 0.06 & 23.09& 220.82 \\
    \hline
    Ours full &  \textbf{0.06} & \textbf{21.81}   & \textbf{89.67}  & \textbf{0.05} & \textbf{23.49}  & \textbf{144.92} \\
    \hline
    \end{tabular}}
    \caption{Ablation studies on the PokingPlants and iPER datasets}
    \label{tab:Ablations}
    \vspace{-4mm}
\end{table}
\section*{Acknowledgements}
This research is funded in part by the German Federal Ministry for Economic Affairs and Energy within the project “KI-Absicherung – Safe AI for automated driving” and by the German Research Foundation (DFG) within project 421703927.

\clearpage

\newpage

{\noindent \Huge \textbf{Appendix}}

\appendix

\section*{Preliminaries}
% \begin{comment}
For all of our reported datasets in the main paper, we provide additional video material resulting in 24 videos in total. For each video three individual cycles are shown (indicated left bottom) as well as the corresponding frame-per-second (FPS) rate (right bottom). Analogous to the main paper, the direction of the poke is indicated by a red arrow starting at the poke location $l$ and the target location is marked by a red dot, if not stated otherwise.
The file structure of the videos is as follows:
\begin{verbatim}
  supplementary_material_244
        |
        +--A-Additional_Visualizations
            |
            +--A1-PokingPlants
            |
            +--A2-iPER
            |
            +--A3-Tai-Chi-HD
            |
            +--A4-Human36M
            |
            +--A5-Qualitative_Comparison
            |
        +--B-Additional_Experiments
            |
            +--B2-Impulse_Model
            |
            +--B3-Generalization
\end{verbatim}
In Section~\ref{sec:add_vis}, we discuss our qualitative results and further emphasize the effectiveness of our approach by visually comparing with two competitors from the main paper. In Section~\ref{sec:exp_add} we acquire two additional experiments which are based on a slightly different training procedure and a different interpretation of the poke: first, we introduce the training procedure. We then visually evaluate the trained model and subsequently show this model to generalize to previously unseen types of plants which are obtained using web-search. After discussing the results, we derive Equation~(3) from the main paper within Section~\ref{sec:deriv}. Moreover, to provide empirical evidence for the effects of selecting the highest order of ODE derivative modeled by our method, we show the benefits of increasing the depth of the RNN hierarchy corresponding to an increase in this order in an additional ablation study in Section~\ref{sec:ode_deriv}. In conclusion we describe implementation and evaluation details in the Sections~\ref{sec:implementation} and ~\ref{sec:eval}.
% and finally discuss the
% Therefore, we first explain the used procedure  followed by a discussion of the resulting generations.
% \end{comment}
% For all of our conducted experiments in the main paper, we here provide additional video material, i.e. 24 synthesized videos in total. These include visualizations of our generated sequences for each reported dataset and also comparisons to two of our competitors for our quantitative experiments, which are discussed in Section~\ref{sec:add_vis}.
% Moreover, we also report two additional experiments, which were conducted for our proposed model with a slightly different training procedure. In Section~\ref{sec:exp_add}, we first explain this procedure, before discussing the resulting synthesized videos for the PokingPlants dataset. Finally we show that our model also generalizes to unseen types of plants obtained from Google image search using the terms 'plant' and 'tree'. Both these results are also contained in the video material, which we highly encourage the reader to view. For each video three individual cycles are shown (indicated left bottom) as well as the corresponding frame-per-second (FPS) rate (right bottom). Analogous to the qualitative results within the main paper, the direction of the poke is indicated by a red arrow starting at the poke location $l$ and the target location is marked by a red dot, if not stated otherwise.
\section{Additional Visualizations}
\label{sec:add_vis}
Subsequently, we discuss our obtained results for each dataset individually, before visually comparing our approach with two competing methods from the quantitative evaluation section. Within the section-specific directory '\verb+...--A-Additional_Visualizations+', each subsection matches its corresponding folder (e.g. 'A.1.PokingPlants' corresponds to '\verb+...--A1-PokingPlants+') containing the discussed video sequences. Each video file compares synthesized sequences of our model for a simulated poke and three pokes initiated by human users with the ground truth sequence starting with the same initial image $x_0$, except for those videos containing the comparisons to the related methods.
\subsection{PokingPlants}
\begin{figure*}
 \begin{center}
 \includegraphics[width=\textwidth]{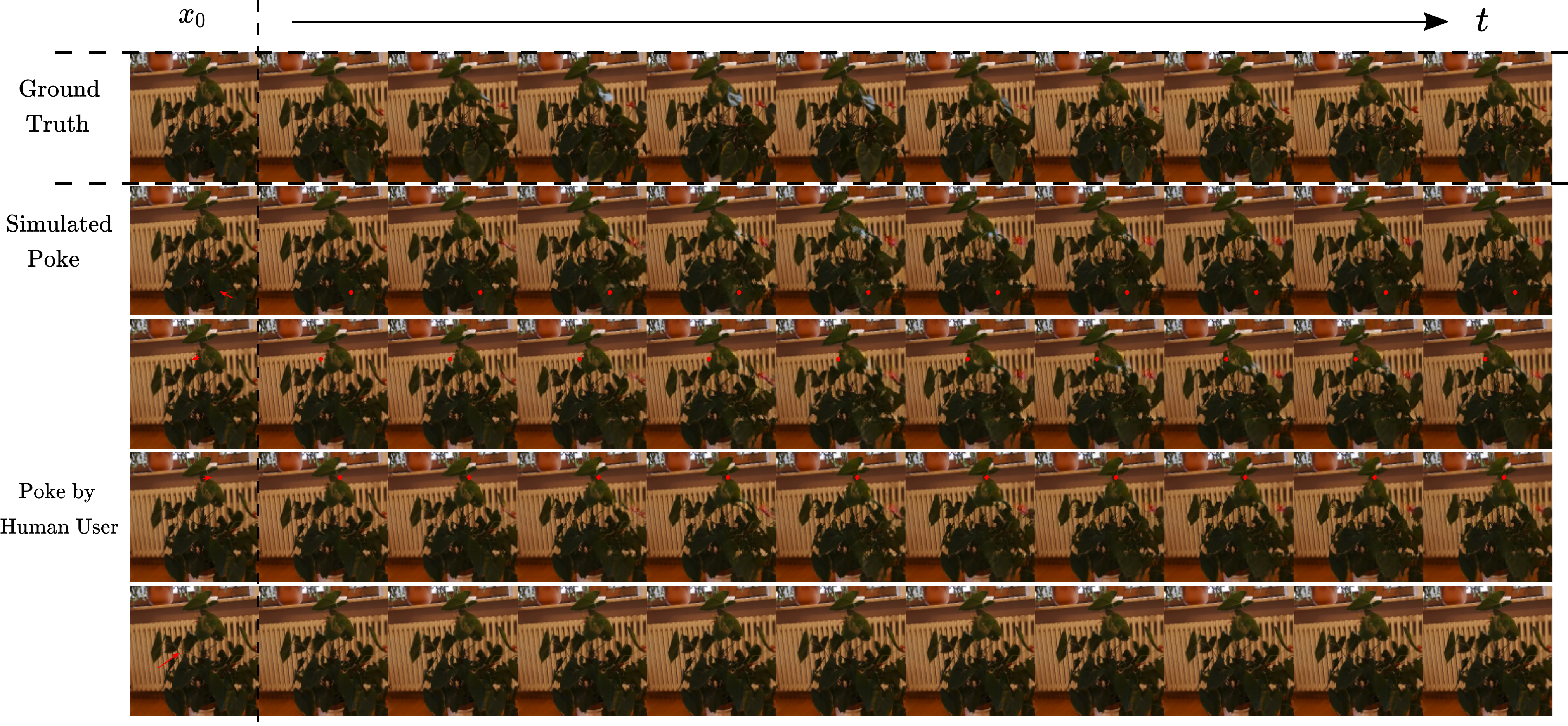}
 \end{center}
 \cprotect\caption{Additional example for the PokingPlants-Dataset: The rows of this Figure depict  the individual frames of the shown sequences in the respective columns of the video example \verb+poking_plants_1.mp4+. The poke direction and magnitude is indicated by the red arrow, which starts at the poke location $l$. The red dot visualizes the target location in each frame.}\label{fig:plants1}
 \end{figure*}

\begin{figure*}
 \begin{center}
 \includegraphics[width=\textwidth]{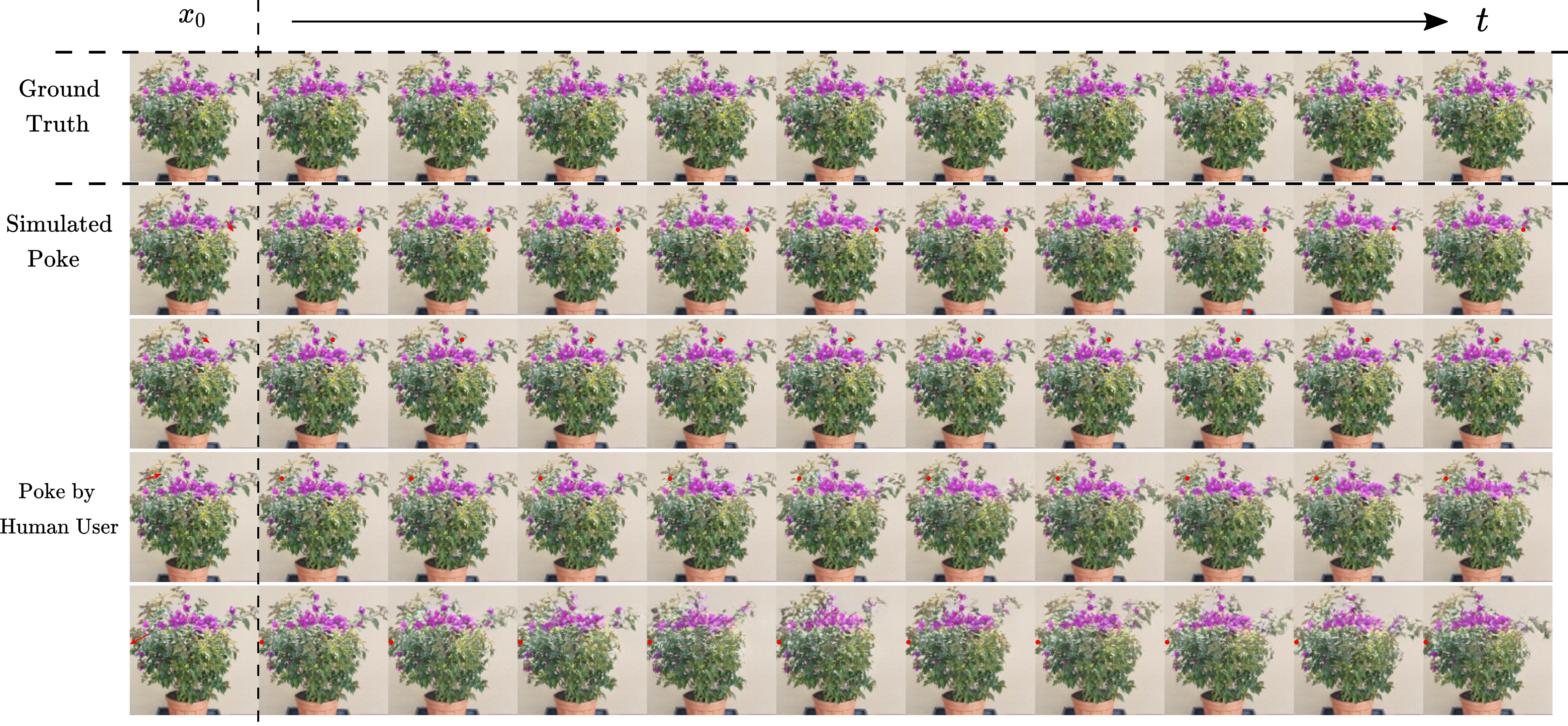}
 \end{center}
 \cprotect\caption{Additional example for the PokingPlants-Dataset: The rows of this Figure depict  the individual frames of the shown sequences in the respective columns of the video example \verb+poking_plants_2.mp4+. The poke direction and magnitude is indicated by the red arrow, which starts at the poke location $l$. The red dot visualizes the target location in each frame.}\label{fig:plants2}
 \end{figure*}

\begin{figure*}
 \begin{center}
 \includegraphics[width=\textwidth]{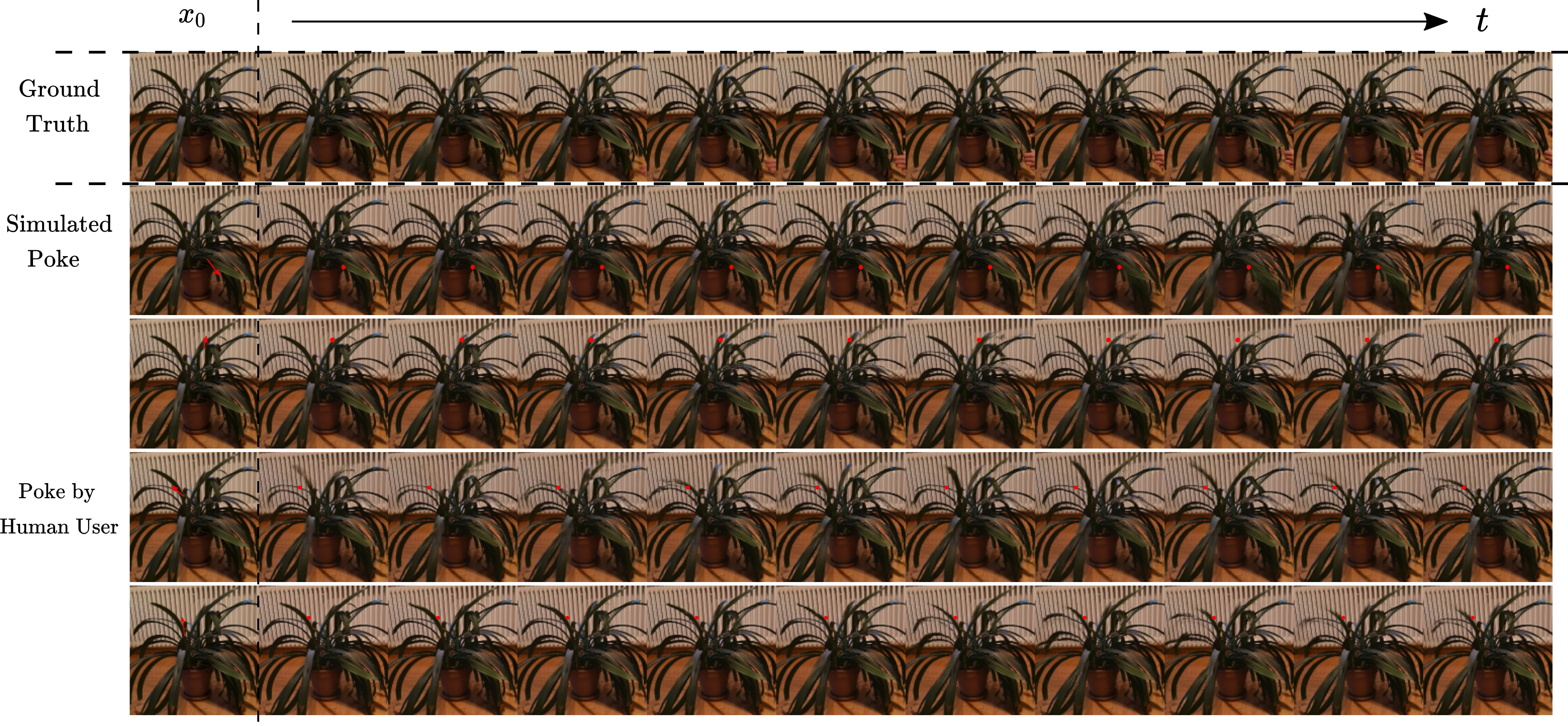}
 \end{center}
 \cprotect\caption{Additional example for the PokingPlants-Dataset: The rows of this Figure depict  the individual frames of the shown sequences in the respective columns of the video example \verb+poking_plants_3.mp4+. The poke direction and magnitude is indicated by the red arrow, which starts at the poke location $l$. The red dot visualizes the target location in each frame.}\label{fig:plants3}
 \end{figure*}

\begin{figure*}
 \begin{center}
 \includegraphics[width=\textwidth]{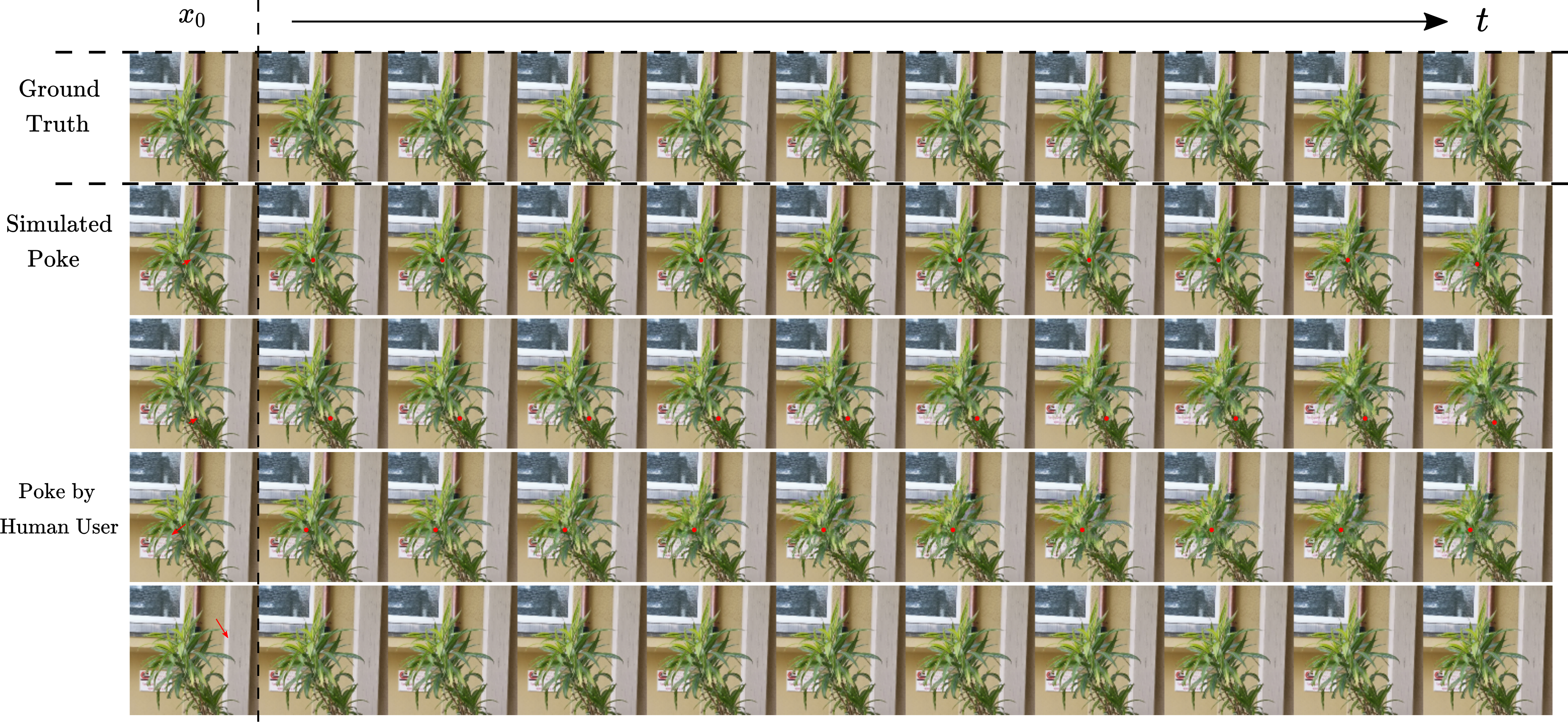}
 \end{center}
 \cprotect\caption{Additional example for the PokingPlants-Dataset: The rows of this Figure depict  the individual frames of the shown sequences in the respective columns of the video example \verb+poking_plants_4.mp4+. The poke direction and magnitude is indicated by the red arrow, which starts at the poke location $l$. The red dot visualizes the target location in each frame.}\label{fig:plants4}
 \end{figure*}

We provide four videos (\verb+poking_plants_[1-4].mp4+) for the PokingPlants dataset showing distinct types of plants of substantially different shapes and appearances. Despite these large variances, our model generates realistic and appealing visualizations which are plausible responses to the poke. Pokes of large magnitudes (indicated by longer arrows) result in larger object motion, affecting not only the object parts in the vicinity of the poke location, but also those parts farther away. This illustrates that our model captures long-term relations between distinct object regions. Modest interactions, in contrast, result in much finer object motions. For instance, the subtle poke initiated in the third column of \verb+poking_plants_2.mp4+ only results in fine-grained movements of those parts next to the poke location, whereas the large interaction in the last column causes nearly the entire plant to heavily oscillate. In summary, the examples demonstrate the capabilities of our method to flexibly model dynamics even for object categories with large intra-class variance, ranging from large scale low frequency motion to very subtle movements. Moreover, our model consistently accomplishes to generate sequences showing the object regions around the poke location to approach the target location. We also provide examples for background interactions in the last columns of the examples \verb+poking_plants_1.mp4+ and \verb+poking_plants_4.mp4+ which indicate our model to separate those pixels comprising the object from background clutter.
Note, that we also plot the individual image frames of the generated videos in Figures~\ref{fig:plants1}-\ref{fig:plants4} together with the target locations and the pokes.
\subsection{iPER}
\begin{figure*}
 \begin{center}
 \includegraphics[width=\textwidth]{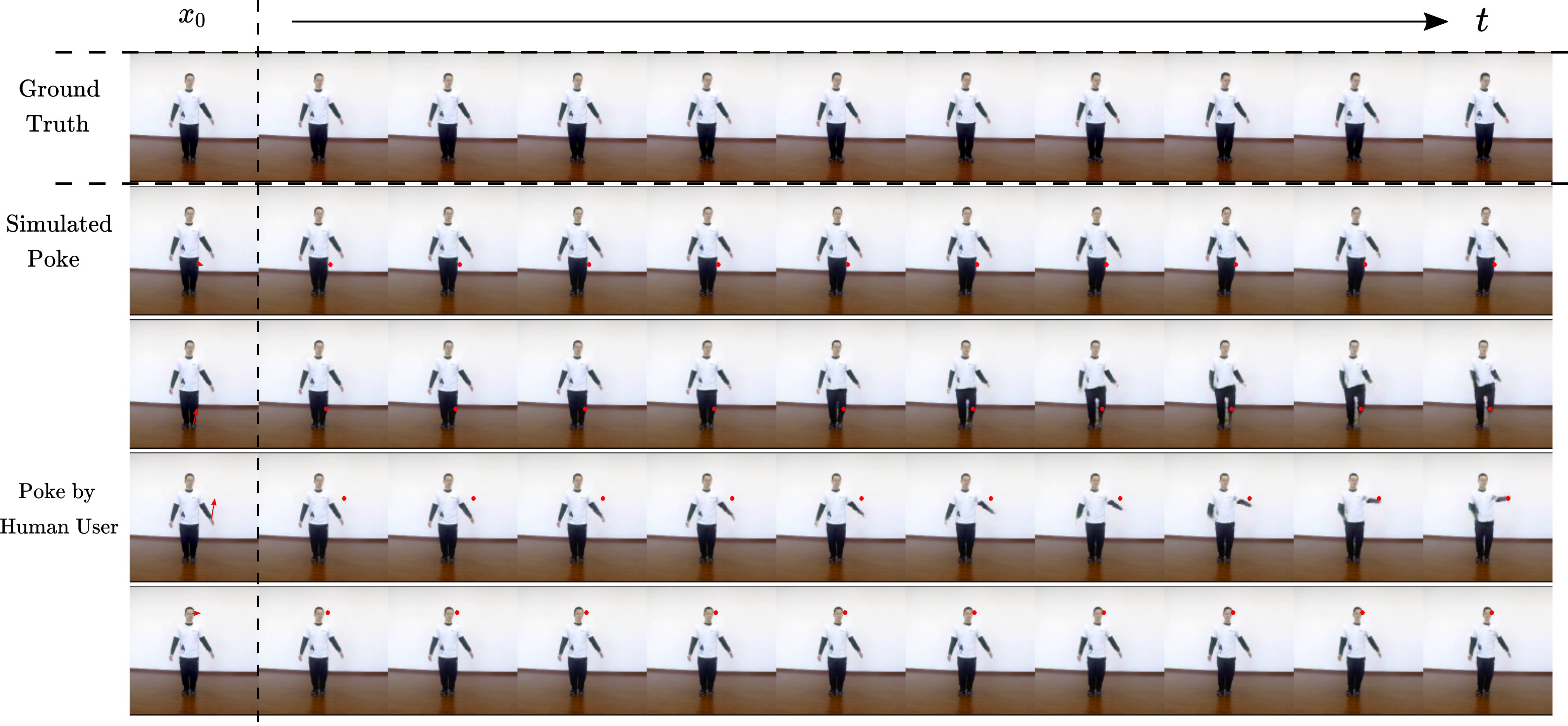}
 \end{center}
 \cprotect\caption{Additional example for the iPER-Dataset: The rows of this Figure depict  the individual frames of the shown sequences in the respective columns of the video example \verb+iper_1.mp4+. The poke direction and magnitude is indicated by the red arrow, which starts at the poke location $l$. The red dot visualizes the target location in each frame.}
 \label{fig:iper1}
 \end{figure*}

\begin{figure*}
 \begin{center}
 \includegraphics[width=\textwidth]{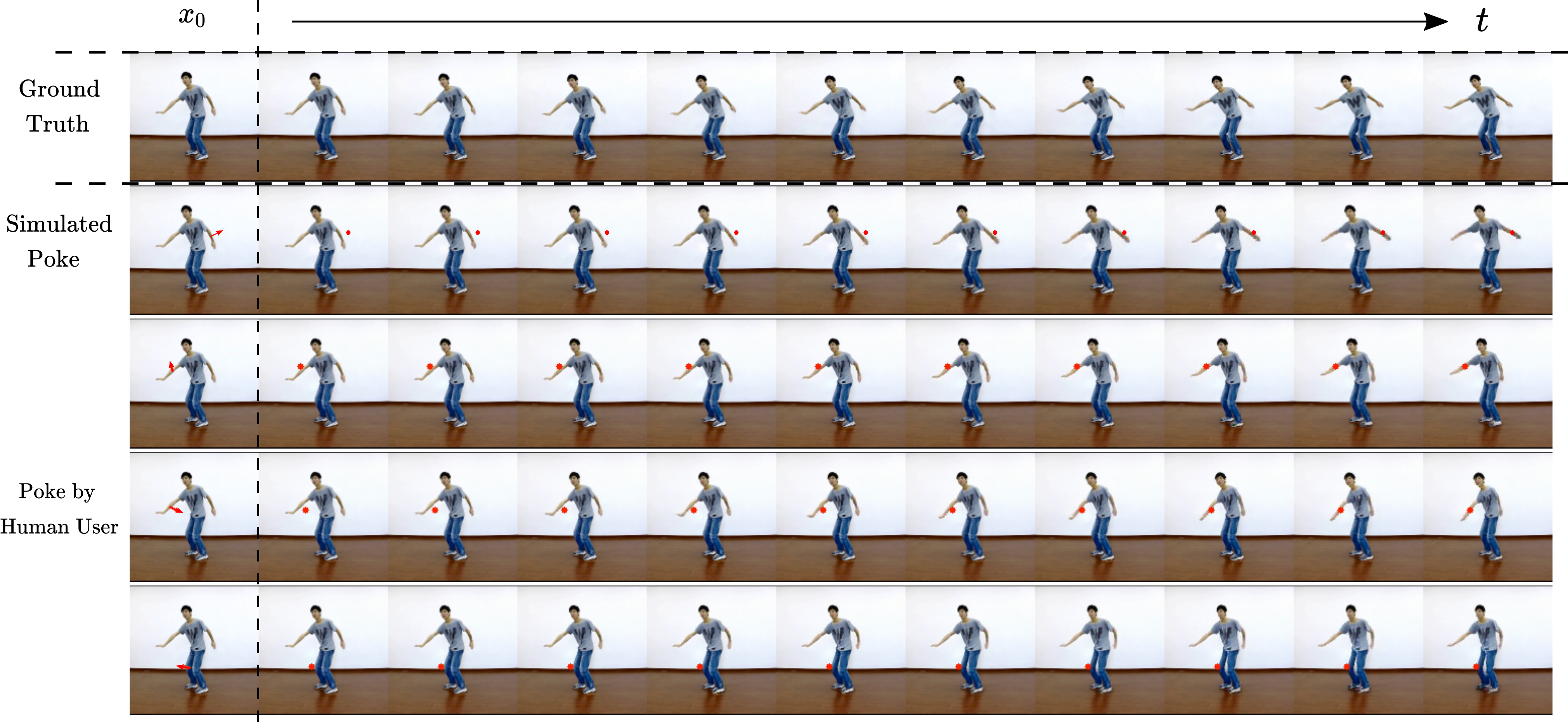}
 \end{center}
 \cprotect\caption{Additional example for the iPER-Dataset: The rows of this Figure depict  the individual frames of the shown sequences in the respective columns of the video example \verb+iper_2.mp4+. The poke direction and magnitude is indicated by the red arrow, which starts at the poke location $l$. The red dot visualizes the target location in each frame.}
 \label{fig:iper2}
 \end{figure*}

\begin{figure*}
 \begin{center}
 \includegraphics[width=\textwidth]{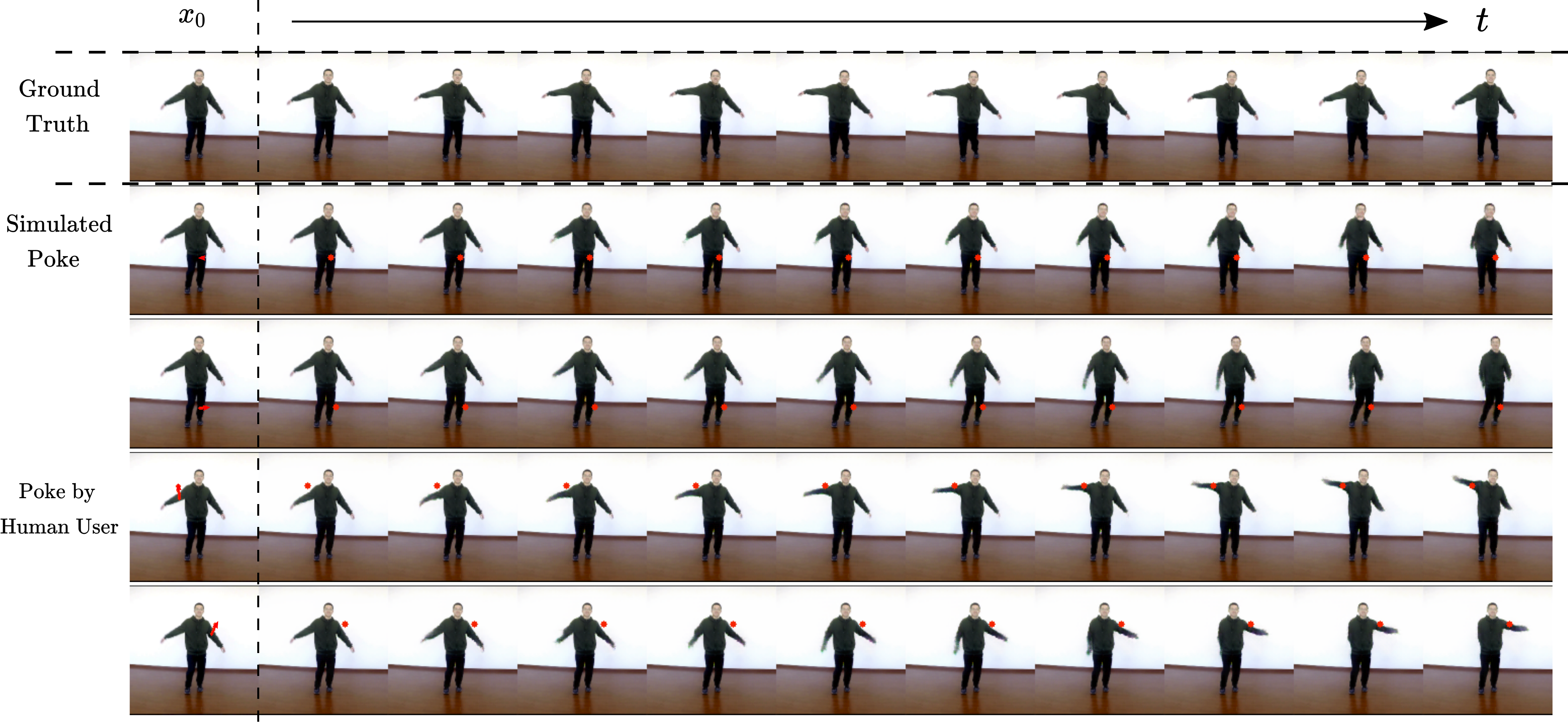}
 \end{center}
 \cprotect\caption{Additional example for the iPER-Dataset: The rows of this Figure depict  the individual frames of the shown sequences in the respective columns of the video example \verb+iper_3.mp4+. The poke direction and magnitude is indicated by the red arrow, which starts at the poke location $l$. The red dot visualizes the target location in each frame.}
 \label{fig:iper3}
 \end{figure*}

\begin{figure*}
 \begin{center}
 \includegraphics[width=\textwidth]{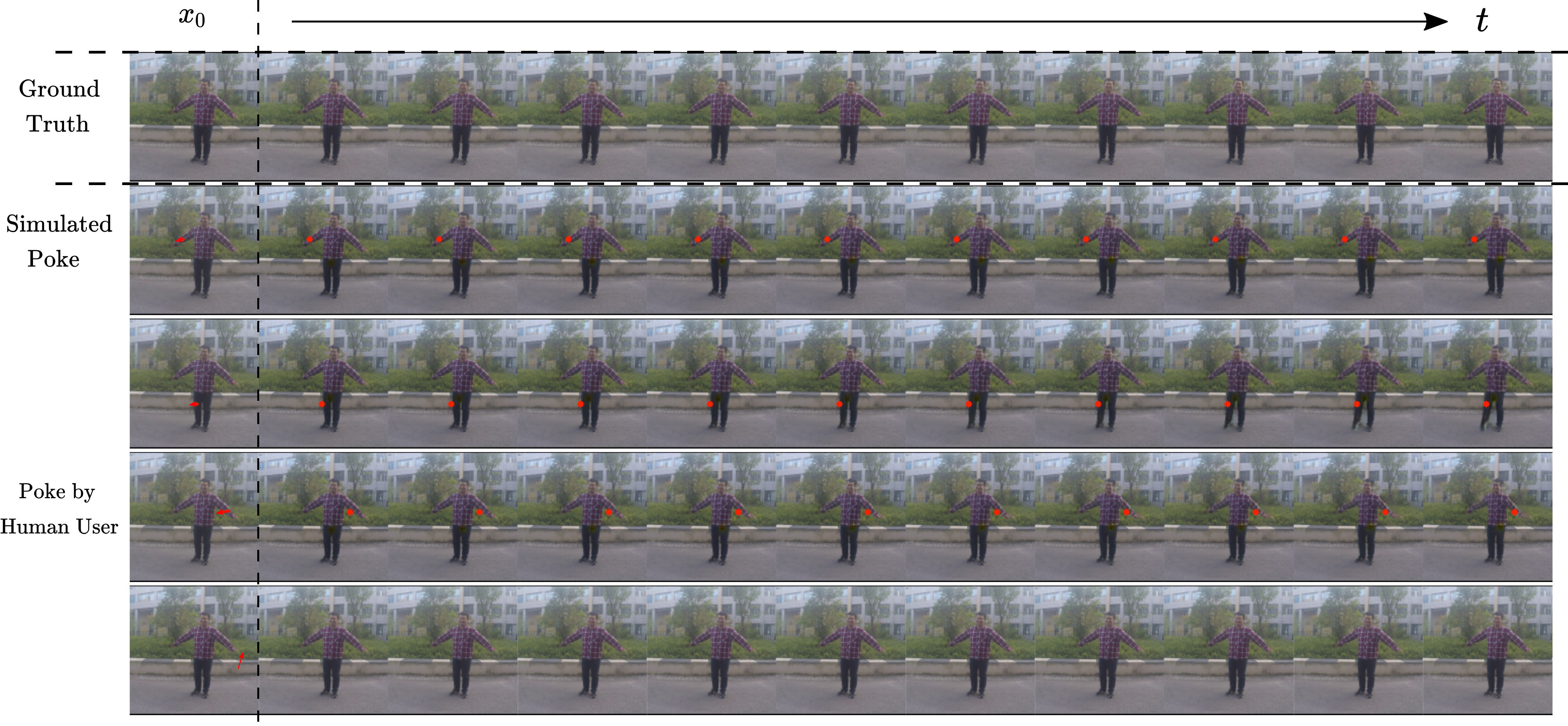}
 \end{center}
 \cprotect\caption{Additional example for the iPER-Dataset: The rows of this Figure depict  the individual frames of the shown sequences in the respective columns of the video example \verb+iper_4.mp4+. The poke direction and magnitude is indicated by the red arrow, which starts at the poke location $l$. The red dot visualizes the target location in each frame.}
 \label{fig:iper4}
 \end{figure*}

For the iPER~\cite{iper} dataset, we also provide four videos (\verb+iper_[1-4].mp4+) containing unseen actors in indoor as well as outdoor scenes. When comparing the synthesized motion, which is generated based on simulated pokes (second columns), with the ground truth sequences depicted in the first columns, one can clearly observe, that our proposed method achieves to infer realistic global motion only from those sparse, localized interactions. The model can even generate complex human dynamics including distinct combinations of the arms and legs resulting from large pokes, as visualized in the third column of \verb+iper_1.mp4+. Furthermore, it generalizes well to out-of-distribution settings as indicated by the example \verb+iper_4.mp4+ showing an actor in the wild~\footnote{Nearly the entire dataset is recorded indoor in front of white background, as indicated by the remaining examples. Therefore, we observe the vast majority of competing models trained on iPER to be struggling with such in-the-wild settings.} and is capable to separate background from foreground also in such cases.
Similar to the PokingPlants dataset, we visualize the individual image frames of the generated sequences together with the pokes and target locations in Figures~\ref{fig:iper1}-\ref{fig:iper4}.
\subsection{Tai-Chi-HD}
\begin{figure*}
 \begin{center}
 \includegraphics[width=\textwidth]{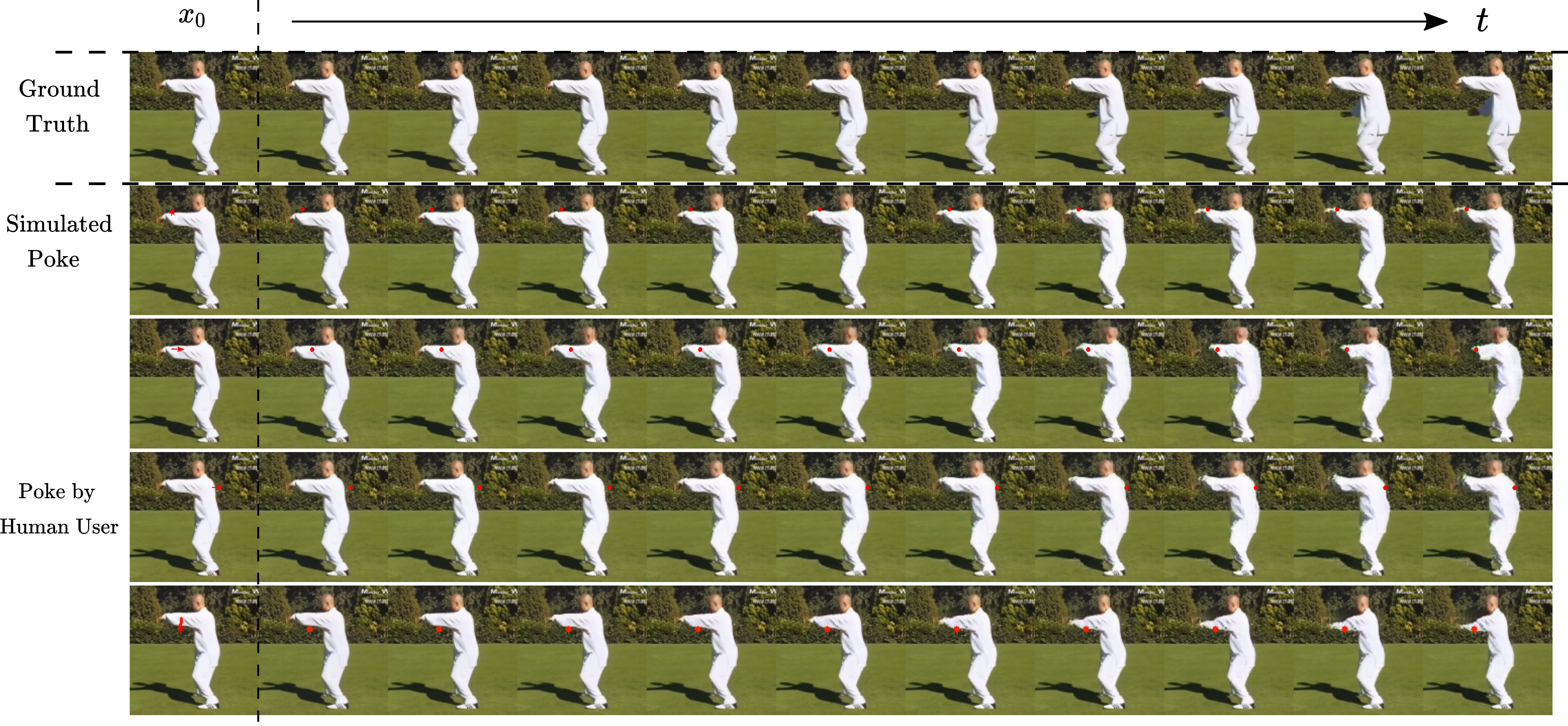}
 \end{center}
 \cprotect\caption{Additional example for the Tai-Chi-HD-Dataset: The rows of this Figure depict  the individual frames of the shown sequences in the respective columns of the video example \verb+taichi_1.mp4+. The poke direction and magnitude is indicated by the red arrow, which starts at the poke location $l$. The red dot visualizes the target location in each frame.}
 \label{fig:taichi1}
 \end{figure*}

\begin{figure*}
 \begin{center}
 \includegraphics[width=\textwidth]{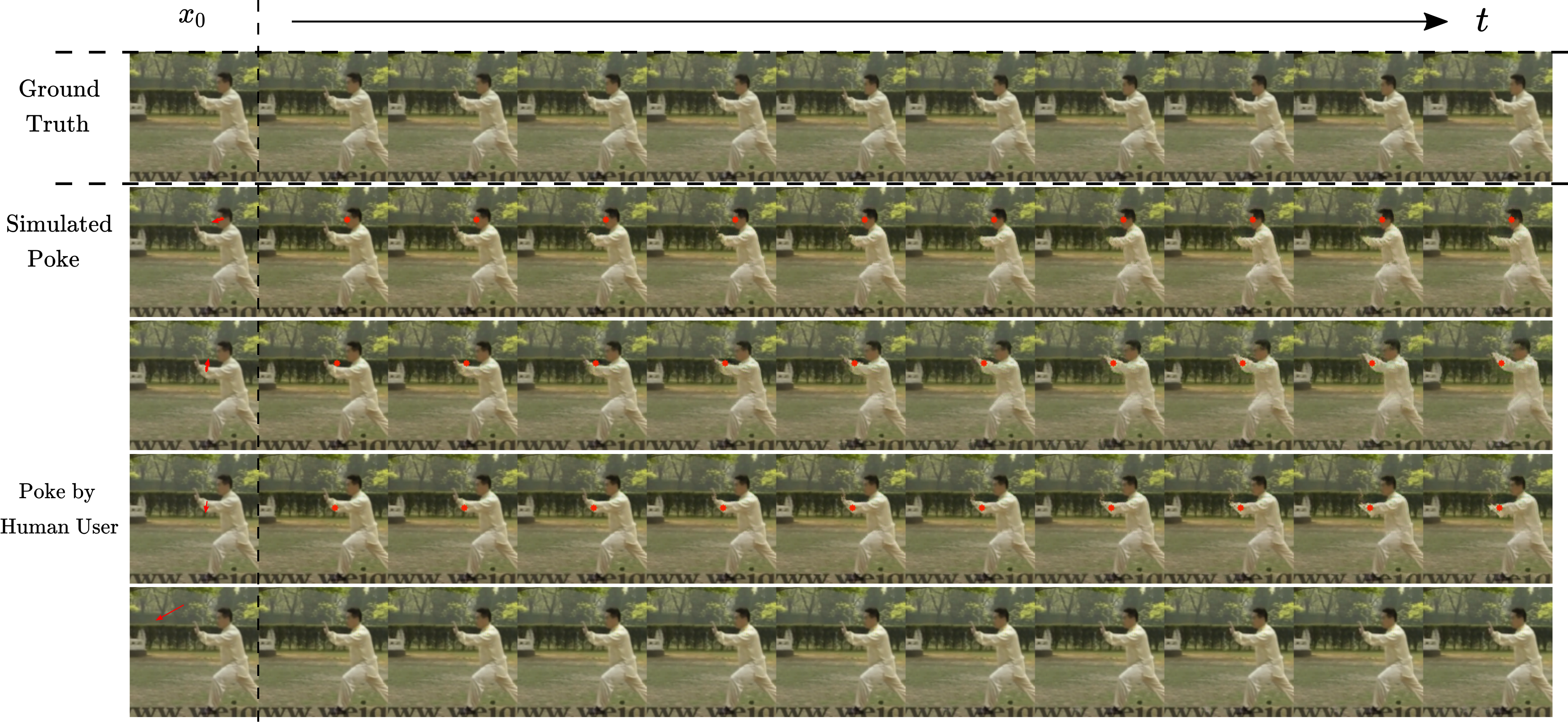}
 \end{center}
 \cprotect\caption{Additional example for the Tai-Chi-HD-Dataset: The rows of this Figure depict  the individual frames of the shown sequences in the respective columns of the video example \verb+taichi_2.mp4+. The poke direction and magnitude is indicated by the red arrow, which starts at the poke location $l$. The red dot visualizes the target location in each frame.}
 \label{fig:taichi2}
 \end{figure*}

\begin{figure*}
 \begin{center}
 \includegraphics[width=\textwidth]{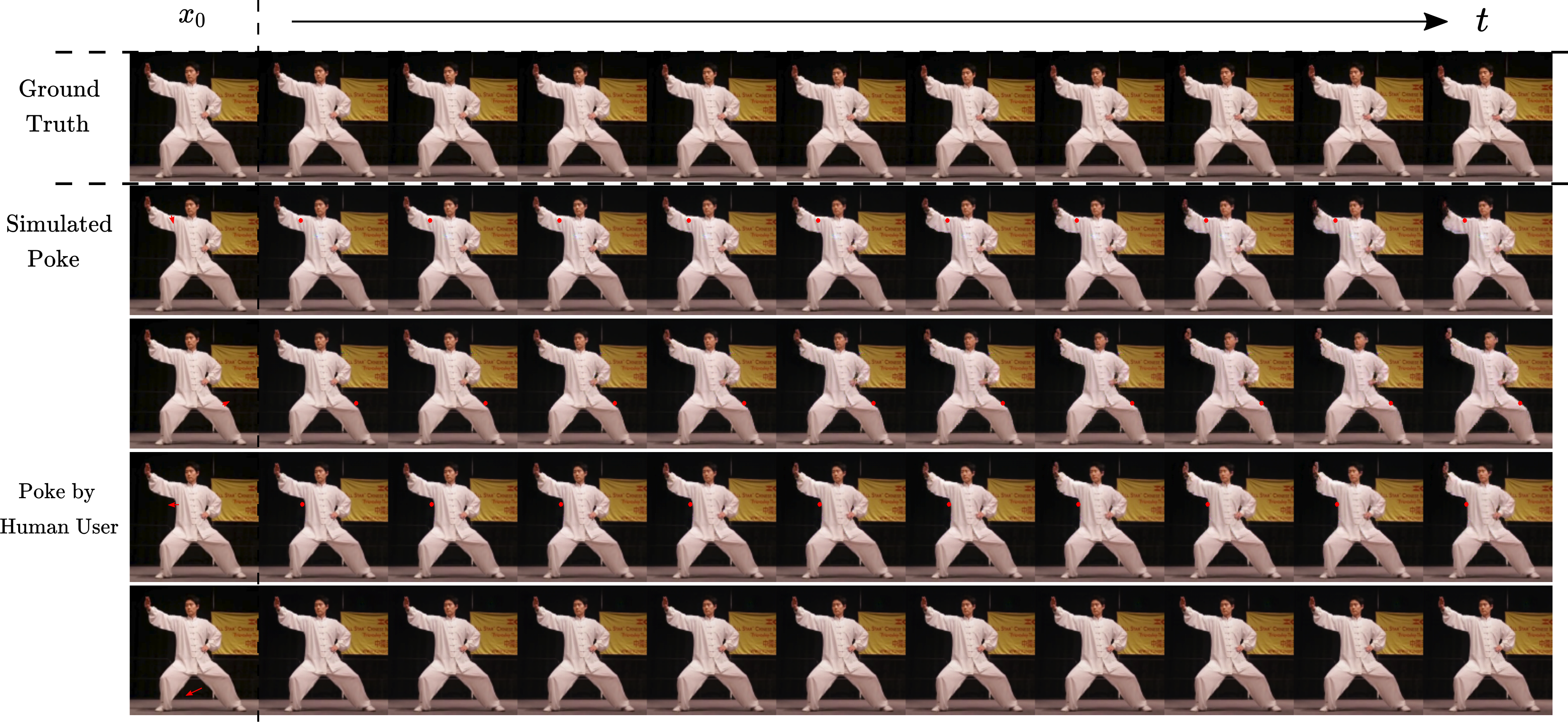}
 \end{center}
 \cprotect\caption{Additional example for the Tai-Chi-HD-Dataset: The rows of this Figure depict  the individual frames of the shown sequences in the respective columns of the video example \verb+taichi_3.mp4+. The poke direction and magnitude is indicated by the red arrow, which starts at the poke location $l$. The red dot visualizes the target location in each frame.}
 \label{fig:taichi3}
 \end{figure*}

\begin{figure*}
 \begin{center}
 \includegraphics[width=\textwidth]{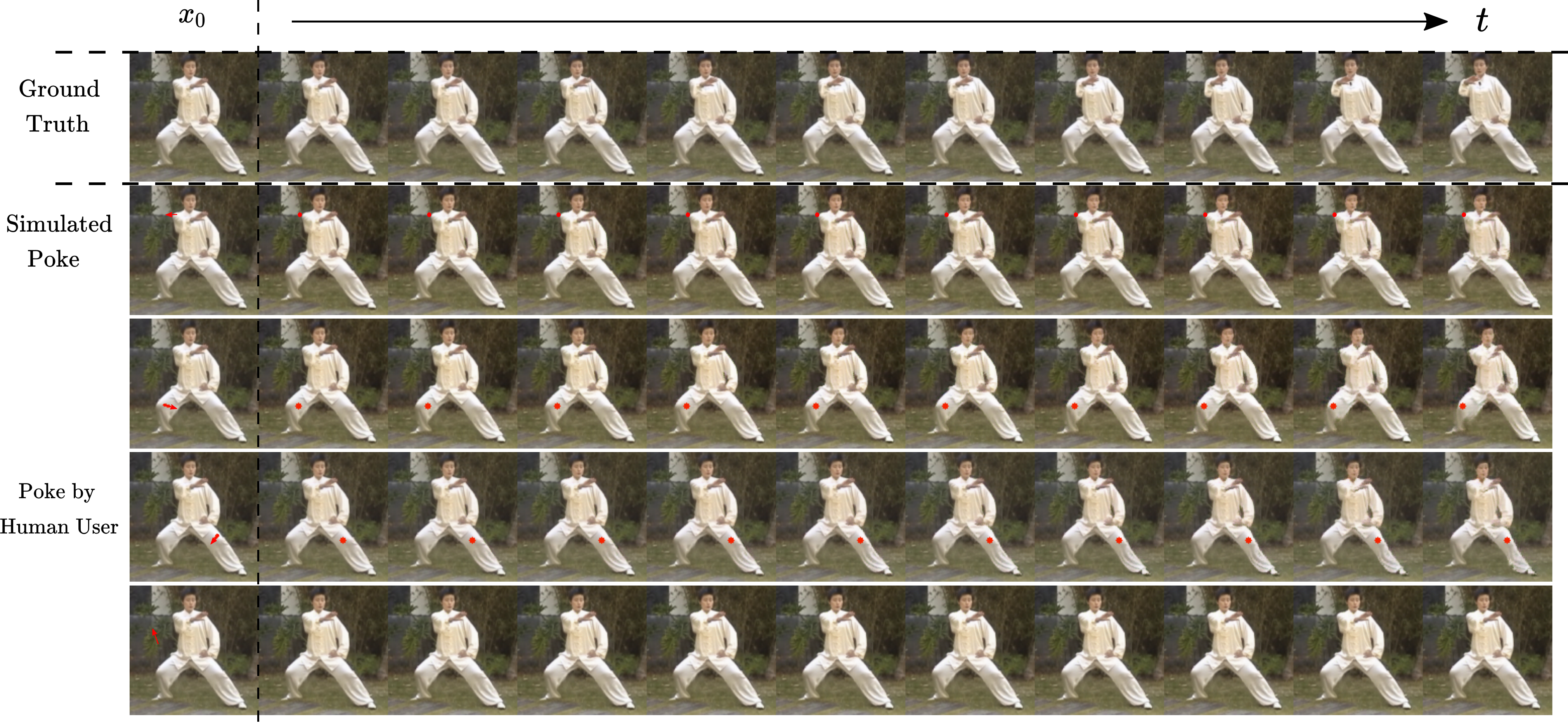}
 \end{center}
 \cprotect\caption{Additional example for the Tai-Chi-HD-Dataset: The rows of this Figure depict  the individual frames of the shown sequences in the respective columns of the video example \verb+taichi_4.mp4+. The poke direction and magnitude is indicated by the red arrow, which starts at the poke location $l$. The red dot visualizes the target location in each frame.}
 \label{fig:taichi4}
 \end{figure*}

We now visualize examples for the Tai-Chi-HD~\cite{first_order_2019} dataset, which does not contain movements as large as those in iPER, but is nonetheless challenging, since it contains many in-the-wild scenes with highly textured background and substantial amounts of camera movements. Thus, we use it to demonstrate our method to be also able to handle such real-world conditions. As for the other datasets discussed so far, we prepared four videos (\verb+taichi_[1-4].mp4+; the individual image frames are shown in Figures~\ref{fig:taichi1}-\ref{fig:taichi4}). Also for this dataset, the model  infers plausible global motions from the poke as indicated by the examples obtained from the simulated poke, which are similar to the ground truth sequences for all provided visualizations. Moreover, the model generates plausible responses to user interactions, demonstrating our model to be applicable to outdoor conditions. This is further emphasized by the capability of our model to separate the object area from the background in the presence of camera movements and background clutter.
\subsection{Human3.6M}
\begin{figure*}
 \begin{center}
 \includegraphics[width=\textwidth]{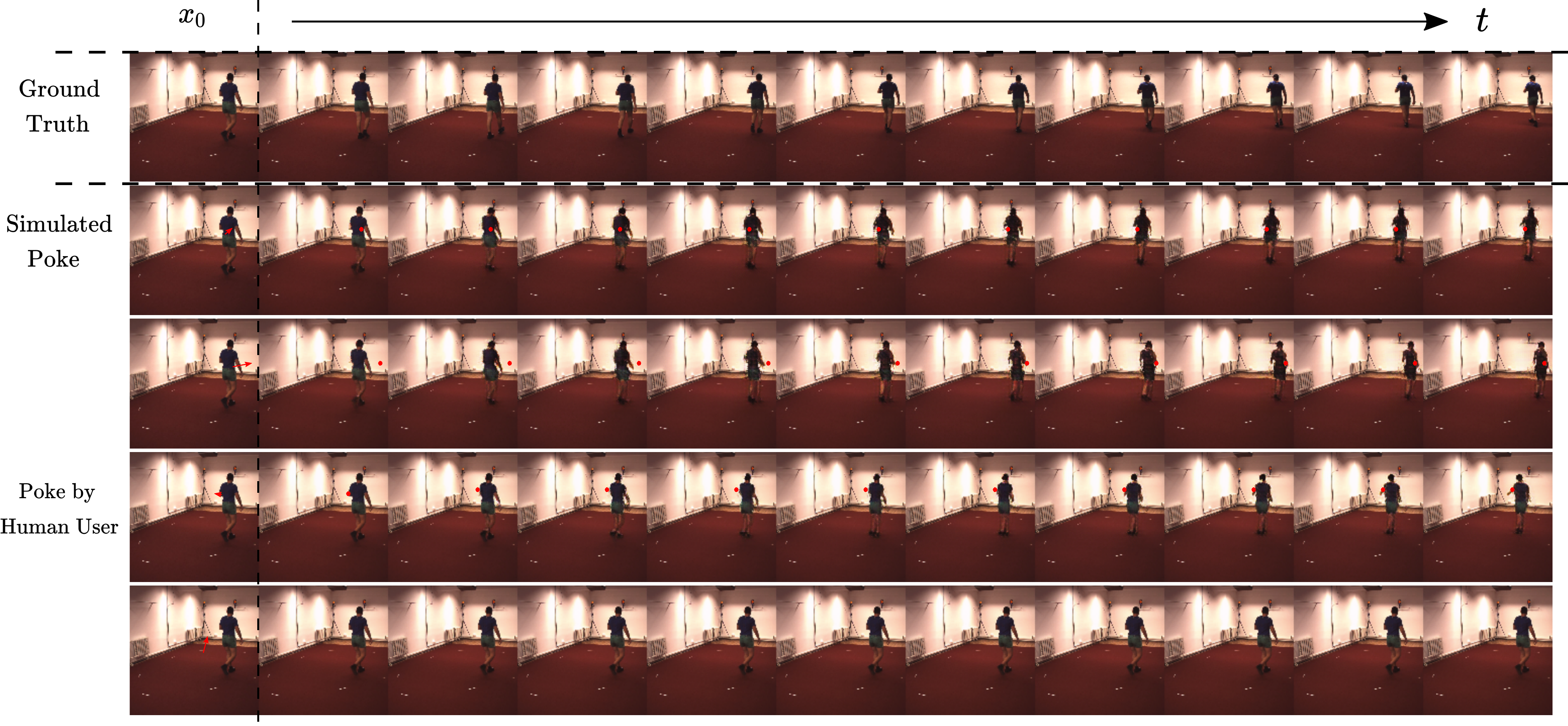}
 \end{center}
 \cprotect\caption{Additional example for the Human3.6M-Dataset: The rows of this Figure depict  the individual frames of the shown sequences in the respective columns of the video example \verb+h36m_1.mp4+. The poke direction and magnitude is indicated by the red arrow, which starts at the poke location $l$. The red dot visualizes the target location in each frame.}
 \label{fig:h36m1}
 \end{figure*}

\begin{figure*}
 \begin{center}
 \includegraphics[width=\textwidth]{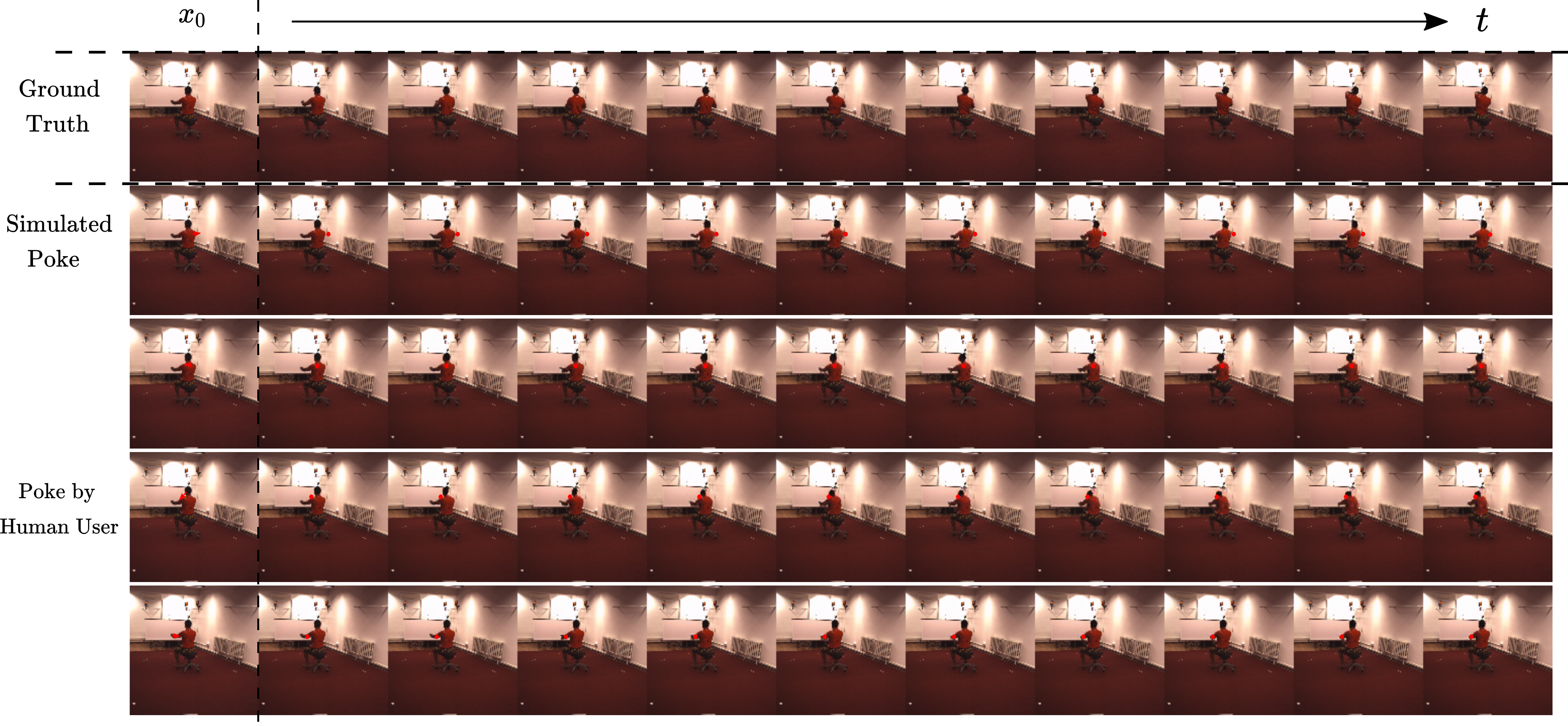}
 \end{center}
 \cprotect\caption{Additional example for the Human3.6M-Dataset: The rows of this Figure depict  the individual frames of the shown sequences in the respective columns of the video example \verb+h36m_2.mp4+. The poke direction and magnitude is indicated by the red arrow, which starts at the poke location $l$. The red dot visualizes the target location in each frame.}
 \label{fig:h36m2}
 \end{figure*}

Finally, we show exemplary results of our method on the Human3.6m~\cite{h36m} dataset, which is challenging due to the complexity of performed motions and the low number of different unique persons. Consequently during testing, appearance swaps after a small number of predicted frames are frequently observed~\cite{2018epva,2019svrnn,Franceschi2020}. As the majority of actions performed by the persons are either based on walking or sitting we visualize examples for both these cases (\verb+h36m_[1,2].mp4+; corresponding image enrollmentin Figure~\ref{fig:h36m1}+\ref{fig:h36m2}). In the first example, we use the poke to control the walking direction of the depicted actor as well as the covered distance. Although changing the appearance, our model generates plausible dynamics which are responses to the individual pokes. In the second example, we interact with the depicted person by manipulating its torso, head and arms. In summary, the visualizations demonstrate that our model is able to synthesize complex human motions and to control their temporal progression based on interactions.
\subsection{Qualitative Comparison}
% \subsection{Qualitative Comparison}
\noindent \textbf{Comparison with SAVP.}
To visually demonstrate the benefits of our proposed model, we compare it to SAVP~\cite{2018savp}, the strongest competitor of all video prediction models considered in this paper. We therefore visualize three sequences showing distinct types of plants from the PokingPlants dataset in \verb+comparison_savp_plants.mp4+. Each row contains a ground truth sequence, followed by our results and those from SAVP~\cite{2018savp}. Especially on this dataset, we observe large differences in the quality of the generated dynamics. We attribute this to the wide range of covered motions for the distinct types of plants within the dataset. Note, that the spatial video resolution for our model and the baseline was chosen to be $64 \times 64$ for these experiments, as this is common practice in video prediction~\cite{2018savp,2019svrnn,vrnn-hier,Franceschi2020}. The visualized sequences indicate that our competitor struggles especially with subtle oscillations corresponding to fine image details. In contrast, our model also successfully synthesizes these fine motion details, thus demonstrating the efficacy of our proposed hierarchical architecture for capturing fine-grained dynamics.
\\
\noindent \textbf{Comparison with Hao et al.}
Finally, we visually compare our model with the controllable image synthesis approach of Hao et al.~\cite{controllable_image} which can also be applied to generate video sequences, as stated in the main paper. The comparisons are conducted on the PokingPlants and iPER datasets and can be viewed in the videos \verb+comparison_hao_plants.mp4+ and \verb+comparison_hao_iper.mp4+. Each video consists of three individual video sequences. Again, we firstly visualize the ground truth on the left, before showing our results and those of the competitor. Moreover, since we aim at comparing the models in terms of visual quality, we do not visualize the pokes and trajectories but only the generated sequences. For both datasets, we can clearly observe that our model significantly improves upon the competitor in terms of motion consistency and visual quality. Especially on iPER, we observe the approach of Hao et al. to be unable to individually move distinct body parts. Instead, they holistically move the entire body resulting in large errors when compared to the ground truth sequence.
\section{Additional Experiments}
\label{sec:exp_add}
Our model formulation also allows for different interpretations of the poke $p$. In this paragraph, we present the alternative interpretation of the poke $p$ as an initial, localized impulse onto the object instead of a local shift of the pixel at location $l$. We first explain how to train the model to achieve this before showing samples of the resulting synthesized videos for a model trained on the PokingPlants dataset. Lastly, we show that the trained model generalizes also well to unseen types of plants obtained from web search.
\subsection{Training Setting}
Recall that in our main experiments, the poke $p$ was defined as the shift between the poke location and its desired target location in the last frame $\hat{x}_T$ of the predicted video sequence $\hat{X}_{1:T}$ (cf. Section 3 in the main paper).\\
Instead, we now normalize the magnitudes within all estimated flow maps to be in $[0,1]$, i.e we remove the information about the exact target location and only retain the direction and a magnitude which does not define a pixel shift anymore. By parameterizing the latent interaction $\phi(t)$ as
\begin{equation}
    \phi(t) = \left\{\begin{array}{ll} \phi, & t = 0 \\
    0, & t > 0 \end{array}\right. \, ,
\label{eq:diff_param}
\end{equation}
with $\phi = E_\phi(p,l)$, $\phi(t)$ can be seen as an initial force to the initial object state. When using this parameterization of $\phi(t)$, we can in fact train our model to produce sequences which do not longer show plausible object reactions to the shift of a pixel. Instead, they now visualize an object response to an initial impulse acting at location $l$. However, as the magnitude of such an initial impulse should influence the amount motion in the reaction of the depicted object, we have to ensure the model to observe sequences with small amounts of motion for pokes with small magnitude and videos showing much motion for these pokes with large magnitude during training. To this end, we estimate the average of motion within each training sequence $X_{0:T}$ as
\begin{equation}
% \resizebox{.5\textwidth}{!}{
    M(X_{0:T}) = \frac{1}{T}\sum_{i=1}^{T} \mathrm{mag}(D(x_i,x_{i-1})),
    % }
\end{equation}
with $D(x_i,x_{i-1}) \in \mathbb{R}^{H \times W \times 2}$ the estimated flow map between two consecutive images $x_i$ and $x_{i-1}$ and $\mathrm{mag}(D(x_i,x_{i-1}))$ the spatial average of flow magnitudes. Using these estimates, we can sample smaller pokes for sequence with smaller amounts of motion and larger ones for those with more motion, resulting in a learned model which indeed synthesizes videos showing object responses to an initial impulse.\\
\subsection{Results}
We show results of our model trained on the PokingPlants dataset using the procedure explained above. We train the model to reconstruct sequences of length $10$ and predict $25$ frames during inference. We here provide 3 unique video examples, which are denoted as \verb+impulse_model_[1-3].mp4+ and are located in the directory \verb+--B-Additional_Experiments+ within \verb+--B1-Impulse_Model+ subfolder. For illustrating the poke within the videos, we now plot a single arrow starting at the poke location. The length of this arrow is proportional to the magnitude of the poke and, thus, defines the amount of motion to be expected within a generated sequence. Within the leftmost examples in the videos \verb+impulse_model_1.mp4+ and \verb+impulse_model_2.mp4+ only a very subtle poke is induced. As our model has learned to couple the poke magnitude to the global amount of motion visualized in the predicted sequence, it generates similarly subtle movements in these cases. Thus, these plants immediately return to their initial states after performing small oscillations.
Pokes with larger magnitudes, however, cause larger motions to elapse. Hence, in these cases, the poked plants do not approach their rest state again within the $25$ frames of the generated video sequence. In summary, the provided videos demonstrate that also for this training setting, the model understands the dynamics of a given object class and, thus, can be used to synthesize video sequences illustrating the response to an initial impulse onto the object body.
% As the target location does not longer exist, we do not plot it within the subsequently presented videos. We only visualize an arrow starting at the poke location. The length of this arrow is proportional to the poke magnitude and consequently defines the amount of motion to be expected within the generated sequence. \\
% We generated three unique video examples for this model, which are denoted as \verb+impulse_model_[1-3].mp4+ and are located in the directory \verb+--B-Additional_Experiments+ within \verb+--B1-Impulse_Model+ subfolder. Our model learns to couple the global amount of generated motion to the poke magnitude. All the three videos contain one example where only a small poke is induced. Consequently, these plants return to their initial states after the first frames. Pokes with larger magnitudes, however, cause larger motions to elapse. Hence, in these cases, the poked plants do not approach their rest state again within the $25$ frames of the generated video sequence.
\subsection{Generalization to Unseen Types of Plants}
Finally, we illustrate the generalization capabilities of our approach by applying it to four unseen types of plants, which were obtained by using image search in the internet. The resulting generations can be viewed in \verb+generalization_[1-4].mp4+ within the subfolder \verb+--B2-Generalization+. The applied model was trained on the PokingPlants dataset as well as the vegetation samples from the Dynamic Texture Database~\cite{dyntex}. For each example, we provide video sequences based on four different pokes and also visualize the nearest neighbor from the train set (last column). The nearest-neighbor is computed in the feature space of our object encoder $\mathcal{E}_\phi$.
For the trees shown in \verb+generalization_1.mp4+ and \verb+generalization_3.mp4+ as well as for the pot plants in the remaining two examples, we observe the model to predict plausible dynamics. Remarkably, the model also captures the relations between distinct parts for the two pot plants, as indicated by the columns two and three of \verb+generalization_2.mp4+. In the second column, where a subtle interaction is applied, only the directly affected lead and some related parts are slightly moving. However, when initiating a poke with larger magnitude, the entire plant is heavily shaking. Finally our model also achieves to separate foreground from background in example \verb+generalization_3.mp4+, despite its substantially textured background area.
\section{Derivation of Equation (3)}
\label{sec:deriv}
In the following section, we will derive our RNN-hierarchy introduced in Eq.~\eqref{eq:hier_n_pre} in the main paper. Prior to the derivation, we will shortly introduce the correspondence between common RNN architectures~\cite{lstm,gru} and the 2-stage Runge-Kutta-Method~\cite{kutta}, an approximation method for solving ODEs. For more details and the exact proof, see~\cite{niu2019recurrent}.

\subsection{Correspondence between RNNs and the 2-stage Runge-Kutta-Method}
Runge-Kutta-Methods~\cite{kutta} are a family of multi-stage, discrete approximation methods for solving ODEs of the form
\begin{equation}
    \Dot{\sigma} = f(\sigma(t))\,.
\label{eq:ode}
\end{equation}
Niu et al.~\cite{niu2019recurrent} showed that common RNN architectures as LSTMs~\cite{lstm} and GRUs~\cite{gru} can be seen as realizations of the 2-stage Runge-Kutta-Method~\cite{kutta} (RK2). Given an initial value $\sigma(0) = \sigma_0$, RK2 approximates $\sigma(t)$ at discrete time-steps $i \in [1,T]$. Let $h$ be the step-size between two consecutive states $\sigma_{i}$ and $\sigma_{i+1}$, which the ODE is approximated at, then the RK2 method reads
\begin{equation}
    \begin{aligned}
        \sigma_{i+1} &=  \sigma_i + h \cdot K_1(\sigma_i) + h \cdot K_2(\sigma_i + h) \\
            &= \sigma_i + K_1^*(\sigma_i) + K_2^*(\sigma_i) \, ,
    \end{aligned}
\label{eq:rk2}
\end{equation}
with $K_1^*(\sigma_i) := h \cdot K_1(\sigma_i)$, $K_2^*(\sigma_i) := h \cdot K_2(\sigma_i + h)$.~\footnote{For the exact schemes to compute $K_1$ and $K_2$, see~\cite{kutta}.}\\
Based on an initial hidden state $\sigma_0$, the update rule of the GRU-cell~\cite{gru} is
\begin{equation}
\begin{aligned}
    \sigma_{i+1} &= \sigma_i + z(\sigma_i) \odot \sigma_i + z(\sigma_i) \odot h(\sigma_i)\\
    &= \sigma_i + G_1(\sigma_i) + G_2(\sigma_i)\, ,
\end{aligned}
\label{eq:gru_update}
\end{equation}
with $z(\sigma_i)$ the update gate vector and $h(\sigma_i)$ the candidate activation~\cite{gru,gru_form} and $\odot$ the hadamard product~\footnote{Note that $G_1$ and $G_2$ are also functions of the input state of the GRU cell~\cite{gru,gru_form}. We will later make use of this.}. By comparing the Equations \eqref{eq:rk2} and \eqref{eq:gru_update}, the correspondence becomes evident. Note that this is only for clarification, for the exact proof, see Niu et al~\cite{niu2019recurrent}.
\subsection{Derivation of RNN-Hierarchy}
We will now use RK2 to solve an $N$-th ODE $f = f(\sigma^{(1)},\sigma^{(1)},...,\sigma^{(N)})$, with $\sigma^{(n)}$ the $n$-th time derivative of $\sigma(t)$. By introducing the following hierarchy of variables~\cite{hale2009ordinary}
\begin{equation}
\boldsymbol{\sigma} =
\begin{pmatrix}
\sigma^0 \\ \sigma^1 \\ \sigma^2 \\ \vdots \\ \sigma^{N}
\end{pmatrix} :=
\begin{pmatrix}
\sigma \\ \sigma^{(1)} \\ \sigma^{(2)} \\ \vdots \\ f(\sigma^0,...,\sigma^{N-1})
\end{pmatrix}
\, ,
\label{eq:ode_system}
\end{equation}
we can map this $N$-th order ODE onto a system of trivial first order ODEs of the form $\sigma^n = \Dot{\sigma}^{n-1} = f_n(\sigma^{n-1})$. The $n$-th element of $\boldsymbol{\sigma}$ is proportional to the $n$-th time derivative of the original variable $\sigma$. Each of these ODEs has the form of Eq.~\eqref{eq:ode} and - given initial values $\sigma^n_0$ - can be approximated in discrete time by using RK2. Thus, we can apply GRU cells $\mathcal{F}_n$ to solve each ODE individually. However, as the argument of each function $f_n$ is the predecessor $\sigma^{n-1}$ of the variable $\sigma^{n}$, which $f_n$ shall be solved for, we use the approximation $\sigma^{n-1}_{i+1}$ as the input state of $\mathcal{F}_n$, resulting in
\begin{equation}
    \sigma^n_{i+1} = \mathcal{F}_n(\sigma^n_i, \sigma^{n-1}_{i+1})
\end{equation}
for $n \in [1,N]$, which is Eq.~\eqref{eq:hier_n_pre} in the main paper.
\section{Ablation Study: Modelled Order of ODE-Derivative}
\label{sec:ode_deriv}
\begin{table}[t]
    \centering
    \resizebox{.45\textwidth}{!}{\begin{tabular}{l|c|c|c|c|c|c}
    \toprule
    Dataset & \multicolumn{3}{c|}{PP} & \multicolumn{3}{c}{iPER}~\cite{iper} \\
    \hline
    Method & LPIPS $\downarrow$ & PSNR $\uparrow$ & FVD $\downarrow$ & LPIPS $\downarrow$ & PSNR $\uparrow$ & FVD $\downarrow$ \\
    \hline
    $N=1$ & 0.08  & 21.13  & 170.74 & 0.07 & 22.62 & 233.22 \\
    $N=2$ &  0.08 &  21.15 &  138.72 & 0.06 & \textbf{23.12} & 178.52 \\
    \hline
    Ours ($N=3$) &  \textbf{0.06} & \textbf{21.81}   & \textbf{89.67}  & \textbf{0.05} & \textbf{23.11}  & \textbf{144.92} \\
    \hline
    \end{tabular}}
    \caption{Ablation study on modelled order $N$ of derivative of ODE approximation corresponding to the depth of the RNN hierarchy. Since the approximation gets more accurate by including higher orders of derivatives, i.e. increasing the depth of the hierarchy, the model improves in prediction accuracy (LPIPS,PSNR) and dynamics consistency (FVD) for increasing $N$.}
    \label{tab:ab_n}
\end{table}
To further provide empirical evidence on the effects of selecting of the highest order of derivative which is modelled by our proposed method corresponding to the depth $N$ of the RNN-hierarchy, we analyze the effects of varying $N$ in this section.
Thus, $N=1$ corresponds to $\boldsymbol{\mathcal{F}}$ consisting of a single RNN cell in the latent bottleneck between $\mathcal{E}_{\sigma}$ and $\mathcal{G}$ and using no skip connections between these sub-networks while learning $\Omega$. Starting from this baseline, we increase the hierarchy until $N=3$ is reached, which constitutes our proposed model. As each additional RNN cell operates on a higher spatial size than its predecessor, we add a skip connection on the respective spatial level between $\mathcal{E}_{\sigma}$ and $\mathcal{G}$, on which this additional RNN operates while learning $\boldsymbol{\mathcal{F}}$. The resulting three models, which are compared, are all training on videos with a spatial size of $64 \times 64$ on the object categories of humans (on the iPER~\cite{iper} dataset) and plants (on the PP dataset) investigated in the main paper.

Tab.~\ref{tab:ab_n} shows that our proposed model, whose RNN hierarchy is capable of modelling a higher number of derivative than the baselines, obtains lower FVD scores, indicating that complex object dynamics can be more accurately modelled by using a deeper RNN hierarchy. This further arises from comparing the performance of the two baselines, where $N=1$ obtains worse results than $N=2$. Additionally, the increasing $N$ yields an enhanced image quality as highlighted by lowered LPIPS and PSNR scores for raised $N$, which is a further benefit of our proposed hierarchy of RNNs.
\section{Implementation Details}
\label{sec:implementation}
Here we give a detailed explanation of the network architecture as well as the training procedure for our model and the baselines which are used for comparison.
\subsection{Network Details}
\noindent \textbf{Encoders} Within the encoders $\mathcal{E}_\sigma$ and $\mathcal{E}_\phi$ we subsequently apply ELU-activated~\cite{elu} 2D-convolutional-layers with kernel size $3$ and stride $2$, until a spatial resolution of $8 \times 8$ is reached, resulting in $N=3$ and $N=4$ layers for video sequences of spatial resolutions $64 \times 64$ and $128 \times 128$. After each conv-layer, we employ instance normalization~\cite{instance_norm}. The last conv-layer is followed by a final ResNet-Block~\cite{resnet}. The initial number of channels is $32$ after the first conv-layers and is increased by a factor of two after each subsequent layer.
\\
\noindent \textbf{Decoder}
The decoder consists $N$ subsequently applied ResNet-Blocks~\cite{resnet} where $N=3$ for a spatial video resolution of $64 \times 64$ and $N=4$ for videos of spatial size $128 \times 128$. Similar to the encoders, the conv-layers within each ResNet-Block are ELU-activated and followed by instance-normalization layers. Upsampling is achieved by using transposed convolutions instead of common convolutions as first layers of those ResNet-Blocks.\\
\noindent \textbf{Hierarchical Image-to-Sequence-Model}
Our hierachical image-to-sequence model consists of a hierarchy of $N$ Conv-GRU cells, where $N=3$ for spatial video resolutions of $64 \times 64$ and $N=4$ for videos of size $128 \times 128$. The upsampling layers $\mathcal{U}_n$ are implemented as transposed convolutions. The hidden dimension of each individual GRU-cell is equal to the number of channels of the encoder $\mathcal{E}_\sigma$ at the same spatial level.\\
\noindent \textbf{Discriminators}
The static discriminator $\mathcal{D}_S$ is implemented as a patch discriminator \cite{pixtopix}. For the temporal discriminator $\mathcal{D}_T$ we use a 3D Resnet-18 \cite{resnet}. Within both discriminators instance-normalization~\cite{instance_norm} is used after each layer.
\subsection{Training Details}
\label{sec:model_train}
\noindent \textbf{Our model}

\textit{Foreground-Background-Separation}
We assume parts of the background of the videos within the train set to be static and the foreground to obtain a sufficient amount of motion, indicated by a specific magnitude of optical flow. Thus, during training, we only consider locations with an optical flow magnitude larger than the mean of magnitudes of the flow map $D$ for sampling the interaction location $l$. However, as we also want our model to separate the pixels on the object surface from those in the background, we sample a tenth of all poke locations in each epoch out of background pixels and construct artificial pokes by sampling the poke magnitudes and angles at these locations from the locations within the foreground. If our model gets such an artificial poke as input, it is trained to reconstruct a still sequence obtained by repeating the source image $x_0$ $T$ times. Thus, the model indeed learns to separate the pixels comprising the object from those in the background, as indicated by our video examples, where we show examples for such artificial pokes for each dataset.

\textit{Discriminators.}
% We explicitly train our model to separate the pixels on the object surface from those in the background.
% and sample also pokes at locations with small optical flow magnitudes during training. Consequently, we assume such pixels to be part of the background area.
The discriminators are optimized using the hinge formulation \cite{lim2017geometric, big_gan_brock}. For stabilizing the GAN training gradient penalty \cite{MeschederICML2018, gulrajani2017improved} is used for the temporal discriminator. Additionally, we add a feature matching loss \cite{high_res_image_syn} to the overall objective $\mathcal{L}$ for $\mathcal{D}_T$, which we weigh with a factor of $2$. For training the spatial discriminator, we sample 16 individual images from the predicted and ground truth sequences.

\textit{Spatial Video of Predicted Videos.}
To conduct the comparison experiments with recent state-of-the-art video prediction models as well as for ablating our method, we trained models on videos of spatial resolution $64 \times 64$. Our models which are compared with the controllable video synthesis method of Hao et al.~\cite{controllable_image} are trained to predict sequences of spatial size $128 \times 128$. All provided videos are outcomes of those models, except for the ones which are visually compared with SAVP~\cite{2018savp} on the PokingPlants-Dataset. In this case, we predicted videos of spatial size $64 \times 64$ to obtain a fair comparison.\\
\noindent \textbf{Video Prediction Models.}
For comparison, we implemented the video prediction baselines~\cite{2018savp,vrnn-hier,Franceschi2020} based on the official provided code from Github~\footnote{https://github.com/edouardelasalles/srvp}\footnote{https://github.com/facebookresearch/improved\_vrnn}\footnote{https://github.com/alexlee-gk/video\_prediction}. As no pretrained models are available for the utilized datasets, we trained models from scratch for all competitors except for SRVP~\cite{Franceschi2020}, which provide a pretrained model for the Human3.6m~\cite{h36m} dataset. All models are trained to predict sequences of length 10 and spatial size $64 \times 64$ based on two context frames. For models trained from scratch, we used the hyperparameters proposed in the respective publications.\\
\noindent \textbf{Controlled Video Synthesis Model.}
The method of Hao et al.~\cite{controllable_image} is implemented based on the official code\footnote{https://github.com/zekunhao1995/ControllableVideoGen} and the provided hyperparameters for all used datasets. We used their proposed procedure to construct the motion trajectories based on the same optical flow which was we used to train our own model. We trained their model to predict images of spatial size $128 \times 128$.
\section{Evaluation Details}
\label{sec:eval}
\noindent \textbf{FVD-Scores.}
To compute the FVD-score~\cite{fvd} for a given model, we generated 1000 video sequences and sampled 1000 random videos of the same length from the ground truth data. Both the real and the generated examples are the input to an I3D~\cite{i3d} model pretrained on the Kinetics~\cite{kinetics} dataset. Subsequently their distributions in the I3D feature space are compared resulting in the reported FVD-scores.\\
\noindent \textbf{Accuracy Metrics.}
All reported accuracy metrics are based on 8000 predicted video sequences and the corresponding ground truth videos. As these metrics are calculated based on individual image frames, we compare each frame of a generated sequence with its corresponding frame ground in the ground truth sequence, resulting in $T$ scores for a predicted video of length $T$, which are subsequently averaged to obtain a scalar value per sequence.

 % % \begin{figure*}
 % % \begin{center}
 % % \includegraphics[width=\textwidth]{figures/plants_id_8322.pdf}
 % % \caption{Additional examples for the PokingPlants-Dataset}
 % % \end{center}
 % % \end{figure*}


 \clearpage

{\small
\bibliographystyle{ieee_fullname}
\bibliography{egbib}
}
\end{document}